\pdfoutput=1
\PassOptionsToPackage{table}{xcolor}
\PassOptionsToPackage{dvipsnames}{xcolor}
\PassOptionsToPackage{pdftex}{xcolor}

\documentclass[11pt]{article}

\usepackage{acl}
\usepackage{times}
\usepackage{latexsym}
\usepackage{subfigure} 
\usepackage{svg}
\usepackage{graphicx} 
\usepackage{float}
\usepackage{appendix}
\usepackage{amsmath}
\usepackage{tcolorbox}
\usepackage{CJKutf8}
\usepackage{multicol}
\usepackage{multirow}
\usepackage{booktabs}
\usepackage{multirow}
\usepackage{makecell}
\usepackage{xspace}
\usepackage{lipsum}
\usepackage{color, colortbl}
\usepackage{caption}
\usepackage{subcaption}

\usepackage{arabtex}
\usepackage{utf8}
\setcode{utf8}
\usepackage{soul,color}

\usepackage{microtype}
\usepackage{graphicx}
\usepackage[pdftex,dvipsnames]{xcolor}

\usepackage{amssymb}
\usepackage{pifont}
\newcommand{\cmark}{\ding{51}}%
\newcommand{\fmark}{\ding{100}}%
\newcommand{\flmark}{\ding{71}}%

\definecolor{olivegreen}{RGB}{107,142,35}
\definecolor{lightolivegreen}{RGB}{157,192,105}

\newcommand{\ours}[0]{Peacock\xspace}
\newcommand{\benchmark}[0]{Henna\xspace}

\newcommand{\jasmine}[0]{AraLLaMA\xspace}

\usepackage[T1]{fontenc}

\usepackage[utf8]{inputenc}
\usepackage{tikzsymbols}
\usepackage{microtype}
\usepackage{xcolor}
\definecolor{mycolor}{RGB}{255, 0, 0} 
\title{
\raisebox{-2.1ex}{\protect\includegraphics[height=4.5\fontcharht\font`\B]{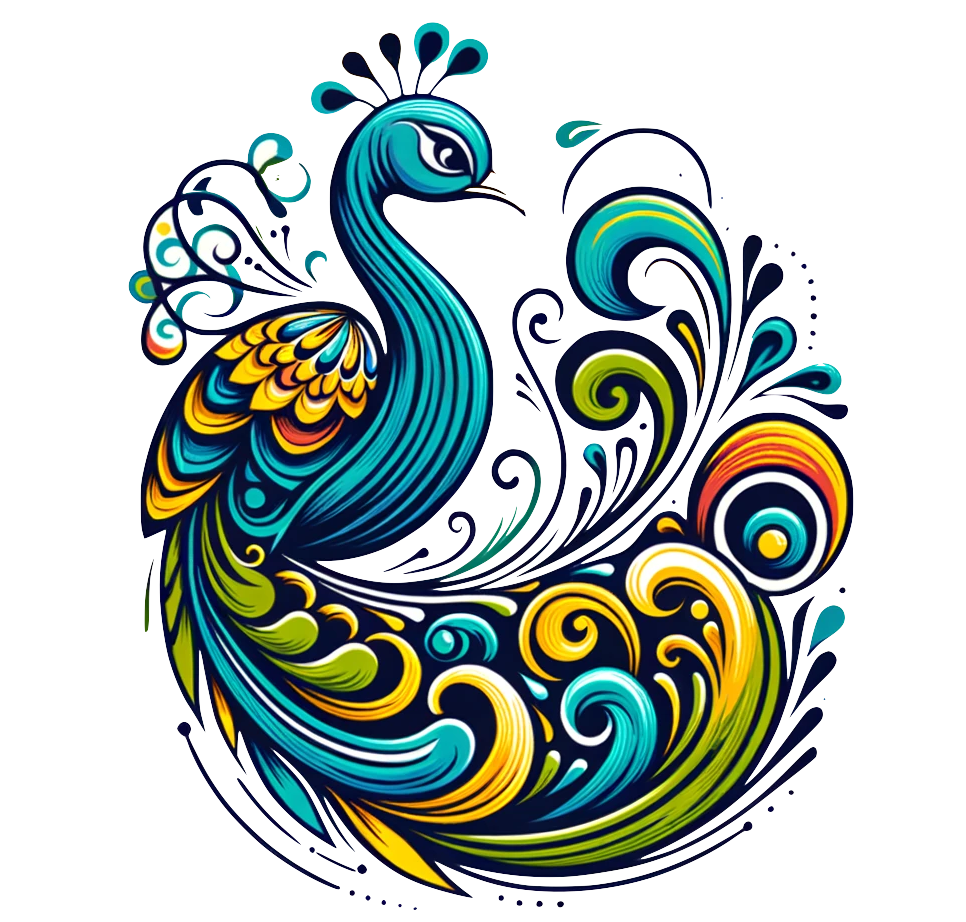}}
\textbf{Peacock}: A Family of Arabic Multimodal Large Language Models and Benchmarks}

\author{
\textbf{Fakhraddin Alwajih}~~~~ \textbf{El Moatez Billah Nagoudi}~~~~ \textbf{Gagan Bhatia}~~~~\\ \textbf{Abdelrahman Mohamed}~~~~ \textbf{Muhammad Abdul-Mageed}~~~~~
\\The University of British Columbia \& Invertible AI \\
\texttt{\normalsize \{muhammad.mageed@\}ubc.ca;invertible.ai} \\  
}

\begin{document}

\maketitle

\begin{abstract}

Multimodal large language models (MLLMs) have proven effective in a wide range of tasks requiring complex reasoning and linguistic comprehension. However, due to a lack of high-quality multimodal resources in languages other than English, success of MLLMs remains relatively limited to English-based settings. This poses significant challenges in developing comparable models for other languages, including even those with large speaker populations such as Arabic. To alleviate this challenge, we introduce a comprehensive family of Arabic MLLMs, dubbed \textit{Peacock}, with strong vision and language capabilities. Through comprehensive qualitative and quantitative analysis, we demonstrate the solid performance of our models on various visual reasoning tasks and further show their emerging dialectal potential. Additionally, we introduce ~\textit{Henna}, a new benchmark specifically designed for assessing MLLMs on aspects related to Arabic culture, setting the first stone for culturally-aware Arabic MLLMs.The GitHub repository for the \textit{Peacock} project is available at \url{https://github.com/UBC-NLP/peacock}.

\end{abstract}

\section{Introduction}

Empowered by progress in large language models (LLMs) and foundation models of other modalities, multimodal large language models (MLLMs) can now have remarkable understanding~\cite{alayrac2022flamingo,li2023blip,dai2023instructblip,liu2023visual,liu2023improved,zhu2023minigpt}. For example, they can handle various complex reasoning tasks spanning from visual question answering to understanding sarcastic comics~\cite{achiam2023gpt,yang2023dawn}. These capabilities, however, are mostly seen in models serving the English language. This leaves behind the majority of the world's languages, furthering an already acute technological divide.  We alleviate this challenge for Arabic, a wide collection of languages and dialects with a native population of more than 400 million speakers.

\begin{figure}[t]
  \centering
  \includegraphics[width=7.5cm]{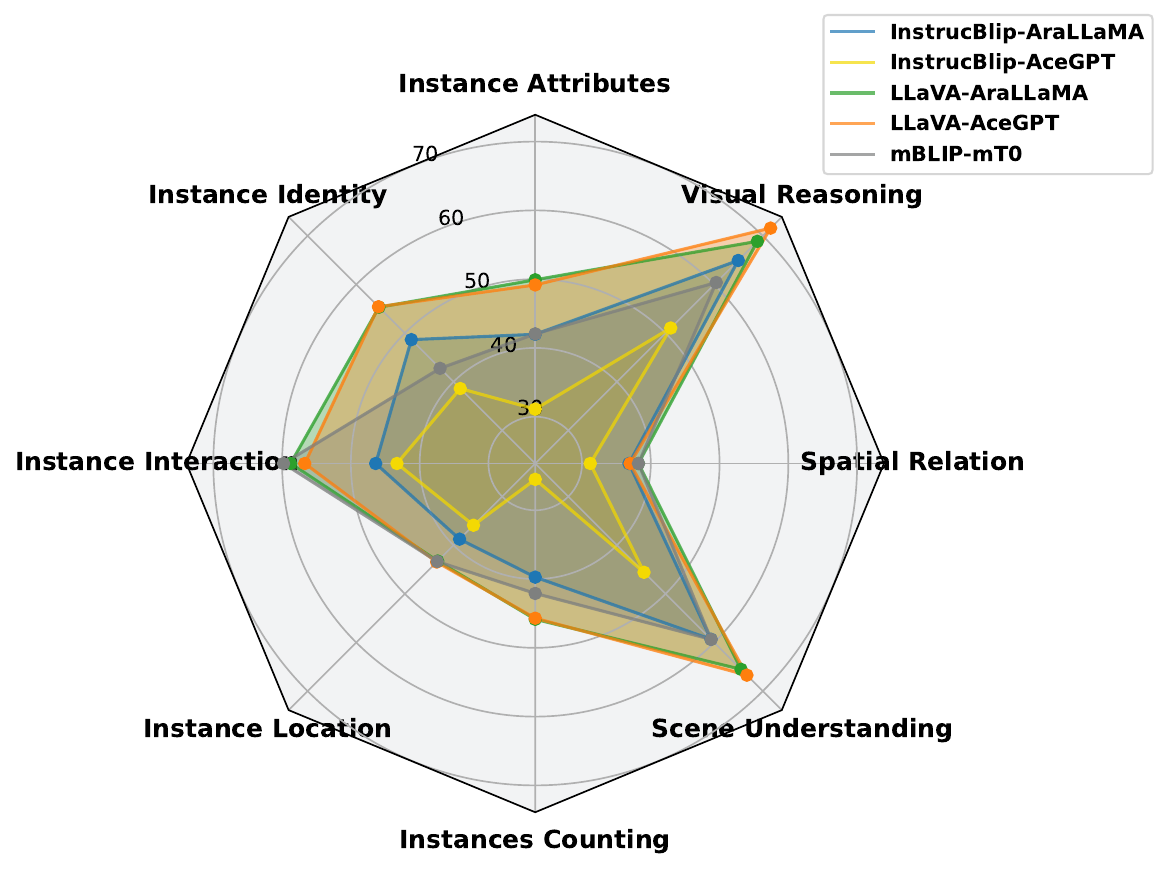}
  \caption{Comparison between the performance of \ours and mBlip models  on SEED-Benchmark dimensions. }
    \label{seed}
\end{figure}

More concretely, we draw inspiration from English counterparts~\cite{dai2023instructblip,liu2023visual} to present a robust family of Arabic MLLMs with powerful vision and language capabilities.
Our models adopt the approach of integrating an image encoder with an Arabic text decoder. In our experimental setup, we explore two popular directions for aligning the vision and the language components: one involves employing a fully connected layer as a projection head on top of the vision encoder~\cite{liu2023visual}, while the other utilizes a Q-former transformer~\cite{dai2023instructblip}.
All models are trained in two stages, a pretraining stage and an instruction finetuning stage. For the first stage, we curate high-quality pretraining data from publicly available English datasets. We translate these datasets into Arabic and apply a carefully designed pipeline to ensure the quality of our training data. Similarly, we curate and translate an instruction finetuning dataset which is essential for achieving reasoning and conversational capabilities.

We showcase the performance of our models across different tasks such as visual question answering (VQA) and visual reasoning. Our models perform much better than a multilingual baseline mBlip~\cite{geigle2023mblip} on different tasks and datasets, and we set the first comprehensive Arabic vision-language benchmark to facilitate future work in this area. 
Finally, we demonstrate the promising capabilities of our \ours models in interacting in dialectal Arabic by conducting a case study on the Egyptian dialect. When finetuned on a small set of Egyptian dialect data, our models exhibit an interesting level of proficiency in their dialectal responses when prompted in the same dialect. We hope this acts as a spark for future works in dialectal Arabic vision language models.

To summarize, our contributions in this paper are as follows:
\textbf{(1)} We introduce a suite of Arabic MLLMs, dubbed \textit{\ours}, capable of instruction following and visual reasoning, in addition to their intriguing dialectal affinity. For developing~\ours, we use existing vision and language models. We also offer a new language model, \textit{AraLLaMA}, based on LLaMA2-7B ~\cite{touvron2023llama}. \textbf{(2)} We introduce a diverse collection of Arabic translated datasets carefully curated for the training and evaluation of Arabic MLLMs. \textbf{(3)} We adapt the popular LLaVA~\cite{liu2023visual} benchmark and SEED-Bench~\cite{li2023seed} for Arabic MLLMs evaluation. \textbf{(4)} We present \textit{\textit{\benchmark}}, a benchmark designed to measure model capabilities in interpreting images related to Arabic culture.

The rest of this paper is organized as follows: In Section~\ref{sec:RW},  we provide an overview of related work. Section~\ref{sec:peacock} introduces Peacock,  our family of MLLMs. In Section~\ref{sec:Benchm}, we describe our evaluation strategies and benchmarks.  In Section~\ref{sec:exp}, we present our experiments, human evaluation,  and a comprehensive analysis of our models. We conclude in Section~\ref{sec:conclusion}.
\section{Related Work}\label{sec:RW}
\subsection{Multimodal Large Language Models}
Progress in MLLMs is largely dependent on advances in LLMs. Refer to Appendix \ref{llm_appndix} for more details on LLM-related works.
The common trend in recent MLLMs involves integrating an LLM as their text decoder alongside a vision encoder for image understanding. Several approaches were proposed for aligning the vision encoder with the text decoder. Flamingo~\cite{alayrac2022flamingo} and Otter~\cite{li2023otter}, for example, blend a \textbf{vision encoder with a resampler and a cross-gated attention layer}, reducing the computational load in vision-text cross-attention and enhancing instruction optimization. While BLIP-2~\cite{li2023blip} and InstructBLIP~\cite{dai2305instructblip}, combine a \textbf{vision encoder with a Q-former and a linear layer}, streamlining the cross-modality projection and utilizing learnable query vectors for feature extraction. LLaVA~\cite{liu2023visual,liu2023improved}, on the other hand, pairs a \textbf{vision encoder with multilayer perceptron (MLP)}, retaining all visual tokens for comprehensive visual information processing. Finally, the simplest form, illustrated by models such as Fuyu~\cite{bavishi2023fuyu8b} and OtterHD~\cite{li2023otterhd}, relies \textbf{solely on a linear layer}, operating as basic decoder-only transformers without specialized vision encoders. This diversity in design showcases the innovative approaches in integrating vision and language in MLLMs.

 \begin{figure*}[!htp]
     \centering
     \begin{minipage}{0.46\linewidth}
         \centering
         \includegraphics[width=7.5cm]{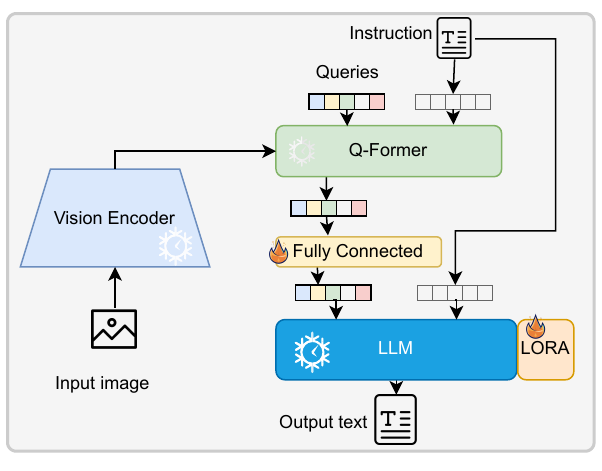}
         \caption{Peacock InstructBLIP architecture: Integrates instruction-specific visual features using Q-Former and a frozen pretrained image encoder.}
         \label{fig:instructblip}
     \end{minipage}
     \hfill
     \begin{minipage}{0.46\linewidth}
         \centering
         \includegraphics[width=7.5cm]{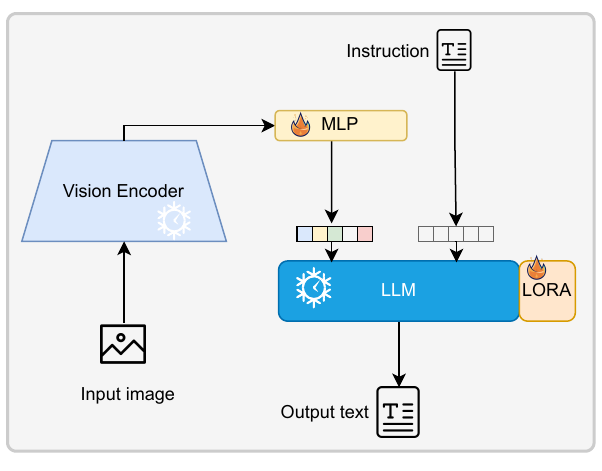}
         \caption{Peacock LLaVA architecture: Combines a pretrained frozen vision encoder with trained Arabic LLMs via an MLP bridge.}
         \label{fig:llava}
     \end{minipage}
\end{figure*}
\subsection{Visual Instruction Tuning}
Following the success of instruction tuning in LLMs, recent works in MLLMs transitioned to visual instruction tuning. MULTIINSTRUCT ~\cite{xu2022multiinstruct} pioneered this transition by creating a multi-modal instruction tuning benchmark dataset that transforms 62 different multi-modal tasks into a unified sequence-to-sequence format. Building on this, LLaVA ~\cite{liu2023visual} leveraged GPT-4's adeptness in understanding multimodal textual representations to reformulate image-text pairs into an instruction-following format. Similarly, MIMIC-IT ~\cite{li2023mimic} focused on generating instruction-response pairs using multi-modal in-context information and a variety of visual scenes. Most recently, M3IT~\cite{li2023m3it} converted traditional vision-language tasks into a unified vision-to-text framework through manual instruction writing and dataset pre-processing. This includes tasks such as captioning, visual question answering, visual conditioned generation, reasoning, and classification. In their comprehensive survey, ~\citet{yin2023survey} provide an extensive overview of MLLMs, including an evaluation of their performance and capabilities. This paper serves as a valuable resource for researchers interested in the field of MLLMs.

\subsection{Arabic MLLMs}
The majority of research in Arabic MLLMs focus mainly on image captioning~\cite{eljundi2020resources,attai2020survey,afyouni2021aracap,emami2022arabic,eddin2022bench,elbedwehy2023improved,mohamed2023violet}. Other areas, VQA for example,  remain largely unexplored. This is primarily due to scarcity of publicly available datasets in these areas. As far as we know, the only significant work in Arabic VQA is by~\citet{kamel2023vaqa} and explores closed-form VQA without attempting generative VQA. 
We also know of no native Arabic datasets for either image captioning or VQA, with two exceptions: AraCOCO~\cite{mohamed2023violet} for image captioning, which is mainly used for evaluation, and AVQA~\cite{kamel2023vaqa} for VQA, which was automatically generated from MSCOCO for Arabic VQA. In many works, translations of either MSCOCO or Flickr8k are utilized~\cite{eljundi2020resources,afyouni2021aracap,mohamed2023violet}. 

\section{\ours} \label{sec:peacock}

\subsection{Architectures}

The \ours family is designed based on the vision components of two architectures, that of InstructBlip~\cite{dai2023instructblip} and LLaVA1.5~\cite{liu2023improved}. For language, our models are integrated with one of two powerful Arabic LLMs, AceGPT~\cite{huang2023acegpt}\footnote{In all our experiments, we use the \textit{AceGPT-7B-chat}. We also limit ourselves to LLMs with 7B parameters due to computational constraints.}  and a new model based on LLaMA2-7B, dubbed~\textit{\jasmine}, that we further pretrain on a large Arabic dataset and finetune using diverse instructions. Our motivation behind introducing~\jasmine is to create a model with strong knowledge of the Arabic language and culture. More information about~\jasmine and how it compares to AceGPT is in Appendix~\ref{jasmine_appndix}.

\noindent\textbf{InstructBlip-Based \ours.} Here, our models consist of four key components: (1) A vision encoder based on the ViT (ViT/G-14) model~\cite{dosovitskiy2020image}, operating at a resolution of 224×224 and employing a patch size of 14. (2) A Querying Transformer (Q-former)~\cite{li2023blip}, designed to link the pretrained vision encoder with the LLM, using the BERT base model~\cite{devlin2018bert} as its foundation. (3) A linear layer projector, tasked with aligning the output of the Q-former with the LLM embedding space. (4) An LLM, incorporating one of the two forenamed models, AceGPT or \jasmine, both of which are derivatives of the LLaMA2 architecture enhanced for Arabic. Figure~\ref{fig:instructblip} illustrates this architecture.

\noindent\textbf{LLaVA-Based \ours.} For this setting, models are structured around three primary components: (1) A vision encoder employing the CLIP-Large model~\cite{radford2021learning}, capable of processing images at a resolution of 336x336 and a patch size of 14, converting these images into 576 tokens. (2) A two-layer MLP projector that aligns the output of the visual and language modalities. (3) And either AceGPT or \jasmine. The architecture is shown in Figure~\ref{fig:llava}

\subsection{Pretraining}
Our models are trained in two stages, a pretraining stage and an instruction finetuning stage. The pretraining stage aims to train the alignment module, which projects the visual and textual features into a common space. The models are trained using our carefully curated text-image pairs dataset. In the case of InstructBlip-based models, only the projection layer, which is the alignment module, is trainable. In contrast, the Q-former, vision encoder and language model parameters are frozen. Meanwhile, for the LLaVA-based models, only the MLP is the trainable part, with the CLIP encoder and LLM being frozen.

\subsection{Visual Instruction Finetuning}
After the pretraining stage, the model will only be capable of generating simple captions and descriptions of an image. To give the models the ability to function on tasks requiring visual reasoning and engage in an intelligible visual conversation, we further finetune them using instruction datasets. 
To keep computational costs manageable, we employ the parameter-efficient finetuning technique LoRA~\cite{hu2021lora}.  Similar to the previous stage, in addition to the LoRA parameters, only the linear layer is trainable in the case of InstructBlip models, while for LLaVA models, we finetune the MLP and apply LoRA to the LLM following the LLaVA 1.5 training scheme~\cite{liu2023improved}.
We provide in Table \ref{tab:parameters} the number of parameters for the main components of each model in Appendix \ref{parameters_appendix}.

\section{Datasets and Benchmarks}\label{sec:Benchm}
\subsection{Translation and Filtering Pipeline}\label{subsec:trans_pipeline}
\label{filtering}
\begin{figure}[ht]
\includegraphics[width=7cm]{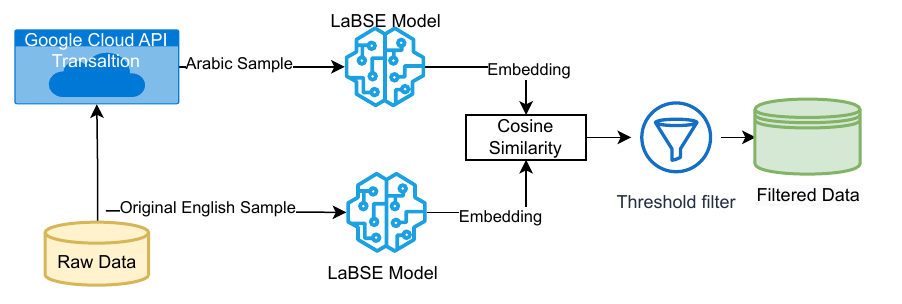}
\centering
\caption{Our data filtering pipeline. After translating the data through Google Cloud API, we obtain the embeddings of both the original and translated samples using the multilingual sentence embedding model LaBSE. For each sample, we calculate the cosine similarity between the two extracted embeddings and reject samples under an 80\% threshold.}
\label{fig:filtering}
\end{figure}

A significant challenge for Arabic MLLMs is lack of available resources, which is due to the difficulty of retrieving relevant Arabic image-text pairs from the internet at scale and absence of suitable image-text relevance filtering methods similar to that of CLIP~\cite{radford2021learning}\footnote{CLIP was used in filtering many English web scraped large-scale datasets~\cite{ordonez2011im2text,sharma2018conceptual,changpinyo2021conceptual}.}. To address this resource gap, we introduce a careful translate-and-filter pipeline for converting publicly available image-text datasets into Arabic without losing data quality. To this end, we adopt the latest version of Google translate API~\cite{GoogleTranslateAPI}. 
We follow ~\citet{mohamed2023violet} in further assuring high quality of acquired translations using a multilingual sentence embedding model LaBSE~\cite{feng2020language}. We calculate the similarity of embeddings between the original and translated sentences (questions and answers), retaining translations that meet a minimum similarity threshold of greater than 80\% for both the question and the answer. Figure~\ref{fig:filtering} demonstrates the filtering pipeline. We provide details about our datasets and translation method in Appendix \ref{app_translation_filtering}, and sample translations illustrating variations in quality ranging from good to moderate to poor in Figure \ref{fig:translation_examples} (also in Appendix \ref{app_translation_filtering}).

\subsection{Pretraining Data}
\label{pretrain_dataset}
Aligning with recent work showing that the quality of LLMs pretraining data is more important than quantity~\cite{gunasekar2023textbooks,lee2023platypus}, we curate a high-quality text-image pairs dataset collected from publicly available sources. Specifically, we utilize  LCS-558K~\cite{liu2023visual} and COCO~\cite{lin2014microsoft} as our pretraining data. LCS-558K encompasses 558k text-image pairs carefully curated from three datasets: LAION~\cite{schuhmann2021laion}, Conceptual Captions~\cite{sharma2018conceptual}, and  SBU~\cite{ordonez2011im2text}. COCO is a high-quality dataset comprising 118k images, covering 80 different objects, with five captions per image, all human-annotated. As stated in Section~\ref{subsec:trans_pipeline}, all the datasets are translated into Arabic using Google API and further filtered based on their semantic similarity with the original English text.

\subsection{Instruction Finetuning Data}
\label{finetuning_dataset}
For the second training stage, we curate another dataset that follows the instructions tuning paradigm as in~\citet{liu2023visual}. Concretely, the model is asked to respond to a specific instruction or question for each image in the dataset. The first dataset we include is the multi-modal instructions dataset by~\citet{liu2023visual}. It comprises 150k samples covering conversations, detailed image descriptions, and complex reasoning instructions and responses. This dataset was created using GPT-4~\cite{achiam2023gpt}, and the images were taken from the COCO dataset. Additionally, we incorporate the VQAv2 dataset~\cite{goyal2017making} after transforming it to the same instructions and responses format. To incorporate further diverse instructions, we utilize 60k multi-choice questions extracted from LLaVA1.5 mixed instruction dataset~\cite{liu2023improved}. This exposes the model to different scenarios, giving it better generalization capabilities. Similar to the pretraining stage, all the datasets are translated using Google API and filtered following our data-cleaning pipeline.

\subsection{Evaluation Benchmarks}

\noindent\textbf{SEED Benchmark.} SEED Benchmark~\cite{li2023seed} (SEED-Bench) consists of 19K multiple-choice questions, each meticulously annotated by humans. These questions cover 12 evaluation dimensions, addressing the comprehension of both image and video modalities. This study exclusively focuses on the image-only subset of SEED-Bench comprising 14K multiple-choice questions. 
SEED-Bench is translated via our translation and filtering pipeline as described in Section \ref{subsec:trans_pipeline}.  

\noindent\textbf{LLaVA Benchmark.} The LLaVA benchmark ~\cite{liu2023visual} (LLaVA-Bench) comprises 30 images, which the authors randomly select from the COCO-Val-2014 dataset. For each image, three questions are generated, resulting in 90 instances. These questions fall into three categories: conversational, detailed description, and complex reasoning. This benchmark evaluates the model’s performance across conversation, description, and reasoning, using GPT-4 scoring.

\noindent\textbf{\benchmark Benchmark.}
As Arabic culture may be underrepresented in current English MLLMs datasets, we develop \benchmark, a new benchmark for testing purposes only. \benchmark comprises attractions, food, events, and other Arabic-relevant objects, consisting of 1,132 samples that have been manually curated and reviewed to ensure quality and relevance. More details about how we create \benchmark and how we use it for evaluation are in Appendix~\ref{ara_bench_appndix}.

\section{Experiments}\label{sec:exp}

\subsection{Implementation Details} 
In the \textbf{first training stage}, we use the 916k image-text pairs described in Section~\ref{pretrain_dataset} to train \ours models. The pretraining phase spans three epochs with a batch size of 32 and a learning rate of 1e-3. As previously described, all model parameters are kept frozen except for the projection layer in the case of InstructBlip-based models and the MLP for LLaVA-based models.
During the \textbf{second training phase}, we utilize the instructions dataset introduced in Section~\ref{finetuning_dataset}.
The models are further finetuned for three epochs with a batch size of eight and a learning rate of 2e-5. As mentioned before, only the introduced LoRA parameters are trainable, with the addition of the projection layer in the case of InstructBlip-based models.

For the training objective, we follow the language modeling approach where the model predicts the next text token given previously predicted text tokens and the visual features. Concretely, our goal is to maximize the probability of the next token or, for mathematical convenience, minimize the negative log-likelihood. The loss is calculated only on the response generated by the model. The instructions and visual tokens are skipped during these calculations.

\subsection{Results and Discussion}

We evaluate our suite of models on a range of typical vision-language tasks and benchmarks. In addition, we show our models' performance on our novel Arabic cultural dataset, \benchmark. Since this is the first work on Arabic MLLMs, we adapt popular benchmarks in the literature to our case. Mainly, these are a VQA-tasks benchmark, LLaVA benchmark~\cite{liu2023visual}, and SEED benchmark~\cite{li2023seed}. We also evaluate the performance on \benchmark benchmark and conduct a case study focusing on the Egyptian dialect. This establishes the first comprehensive benchmark for future works in Arabic MLLMs. We further compare our models with the multilingual mBlip model ~\cite{geigle2023mblip} as a baseline for completeness. The mBlip model is trained on 96 languages, including Arabic.

\subsubsection{General VQA}

\begin{table*}[ht]
\centering

\resizebox{\linewidth}{!}{%
\begin{tabular}{l|l|l|cc|cc|cc}
\toprule
\textbf{Model~~} & \textbf{Architecture~} & \textbf{\multirow{2}{*}{~~~~~~\textbf{LLM}}}  & \multicolumn{2}{c}{\textbf{VQAv2}} & \multicolumn{2}{c}{\textbf{OKVQA}} & \multicolumn{2}{c}{\textbf{GQA}} \\ 
\cline{4-9} 
& & & \textbf{All} & \textbf{Filtered} &  \textbf{All} & \textbf{Filtered} & \textbf{All} & \textbf{Filtered} \\  
\toprule
\multirow{2}{*}{\textbf{Baseline~}} &\multirow{2}{*}{\textbf{mBlip}} & mT0-XL-5B& 38.55& 50.8 & 8.59& 18.18& 35.95&50.45\\
& & BLOOMZ-7B& 41.00& 55.7&11.87& 23.30& 38.55& 54.85\\ 
\midrule
\multirow{2}{*}{\textbf{\includegraphics[scale=0.046]{figs/logo3.png}}} 
&\multirow{2}{*}{\textbf{InstructBlip}}  &   \cellcolor{blue!10} \jasmine& \cellcolor{blue!10} \bf{44.55}& \cellcolor{blue!10}56.15& \cellcolor{blue!10}\bf{20.97}& \cellcolor{blue!10}\bf{29.77}& \cellcolor{blue!10}\bf{42.60}& \cellcolor{blue!10}\bf{58.05}\\

& & \cellcolor{yellow!10} AceGPT& \cellcolor{yellow!10}39.00& \cellcolor{yellow!10}51.20& \cellcolor{yellow!10}10.69& \cellcolor{yellow!10}16.82& \cellcolor{yellow!10}37.00& \cellcolor{yellow!10}57.60 \\
& \multirow{2}{*}{\textbf{LLaVA}} & \cellcolor{green!10} \jasmine& \cellcolor{green!10}40.85& \cellcolor{green!10}52.45& \cellcolor{green!10}14.79& \cellcolor{green!10}25.57& \cellcolor{green!10}33.45& \cellcolor{green!10}49.75\\
& &  \cellcolor{orange!10} AceGPT& \cellcolor{orange!10}41.45& \cellcolor{orange!10}\bf{56.65}& \cellcolor{orange!10}15.14& \cellcolor{orange!10}26.36& \cellcolor{orange!10}33.27& \cellcolor{orange!10}52.20\\
\bottomrule
\end{tabular}
}

\caption{The zero-shot performance of our Peacock models against mBlip on the dev set of different VQA datasets. Models are evaluated on the exact match with the open-generation metric, where an answer is considered correct if it matches any ground truth answers. The baseline is mBlip with different LLMs (mT0-XL-5B and BLOOMZ-7B).} 
\label{tab:general_vqa}
\end{table*}

In general VQA tasks, the challenge involves answering textual questions about images, requiring learning and integrating visual and textual information. This demonstrates a deep understanding of the interconnectedness between the two modalities. To evaluate performance in general VQA, our validation process includes three datasets: VQAv2~\cite{goyal2017making}, OKVQA~\cite{marino2019ok}, and GQA~\cite{hudson2019gqa}. Notably, evaluation of English VQA tasks is typically performed through online platforms by submitting results. However, this option is unavailable for Arabic-translated data because these platforms only support English and the original datasets. Since the test sets do not contain ground truth labels, we evaluate held-out validation sets. We follow~\citet{geigle2023mblip} in using ``exact match accuracy with open generation" to evaluate our models' output. The metric considers an answer correct if it matches any of the ground truth answers. 
Table~\ref{tab:general_vqa} shows model accuracy on these datasets under the zero-shot setting for both the filtered and unfiltered(All) versions of the datasets. 

It is evident from Table~\ref{tab:general_vqa} that the top-performing \ours model, InstructBlip with \jasmine LLM, significantly outperforms the best version of mBlip, which integrates the BLOOMZ-7B LLM, by an average margin of 4.5 points. A comparative analysis of all models also reveals significant performance improvements when only the filtered high-quality data is included. This enhancement is consistently observed across all models and tasks, highlighting the crucial role of data quality in the effectiveness of these models.

Furthermore, we observe that the choice of the underlying LLM is has a significant impact on performance. This is the case if we compare integrating \jasmine to AceGPT in our overall MLLMs. Specifically, the InstructBlip model integrated with \jasmine demonstrates superior performance across all tasks and datasets when using either filtered or unfiltered(All) data in training. This performance disparity is likely attributable to the inherent differences in how these LLMs handle Arabic, with \jasmine being more effective due to its extensive training and its ability to align with visual information. In addition, it is worth noting that the performance of \ours models varies considerably depending on the task, with a general trend of models performing better on the VQAv2 task than on OKVQA and GQA. Such variations can be attributed to each task's inherent complexities and specific requirements, including the sophistication of the presented questions and the nature of the required visual understanding.

\subsubsection{LLaVA Benchmark}
\label{L-bench}

For evaluation using LLaVA benchmark, we follow the method of~\citet{liu2023visual}. Table \ref{tab:llava_bench} displays our models' successful performance across the three metrics of the LLaVA-Bench. Despite the limited data and resources, this suggests a burgeoning capability for multi-modal comprehension in Arabic. Under the same training conditions, the integration of InstructBlip with \jasmine notably excels within the \ours suite. It achieves an average score of 82.27 on the GPT-4 scale, a significant 9.4 margin over the LLaVA model combined with AceGPT. As shown in Table~\ref{tab:llava_bench}, all \ours models surpass the mBlip-BLOOMZ-7B baseline in the three metrics of the LLaVA-Bench.

\begin{table}[t]
\centering
\resizebox{\columnwidth}{!}{%
\begin{tabular}{@{}ll|cccc@{}}
\toprule
   \textbf{Architecture} & \textbf{LLM} & \textbf{Conv} & \textbf{DD} & \textbf{CR} & \textbf{Avg}\\ \midrule
 \textbf{mBlip} & BLOOMZ-7B & \cellcolor{white}55.26 & \cellcolor{white}47.89 & \cellcolor{white}55.43 & \cellcolor{white}52.90\\ \midrule
  \textbf{\multirow{2}{*}{\textbf{InstructBlip}}} & \cellcolor{blue!10} \jasmine & \cellcolor{blue!10}\bf{84.56} & \cellcolor{blue!10}\bf{80.00} & \cellcolor{blue!10}\bf{82.11} & \cellcolor{blue!10}\bf{82.27}\\
 & \cellcolor{yellow!10} AceGPT & \cellcolor{yellow!10}73.28 & \cellcolor{yellow!10}61.40 & \cellcolor{yellow!10}72.67 & \cellcolor{yellow!10}69.13\\
  \textbf{\multirow{2}{*}{\textbf{LLaVA}}} & \cellcolor{green!10} \jasmine & \cellcolor{green!10}75.62 & \cellcolor{green!10}65.01 & \cellcolor{green!10}72.33 & \cellcolor{green!10}71.07\\
 & \cellcolor{orange!10} AceGPT & \cellcolor{orange!10}77.81 & \cellcolor{orange!10}68.85 & \cellcolor{orange!10}73.89 & \cellcolor{orange!10}73.61\\
  \bottomrule
\end{tabular}
}

\caption{Performance of Peacock models and mBlip on LLaVA-Bench scored by GPT-4. Conv: Conversation. DD: Details Description. CR: Complex Reasoning.}
\label{tab:llava_bench}
\end{table}

\subsubsection{SEED Benchmark}

\begin{table}[t]
\centering
\resizebox{\columnwidth}{!}{%
\begin{tabular}{@{}ll|cccccccc}
\toprule
\textbf{Architecture} & \textbf{LLM} & \textbf{IA} & \textbf{II} & \textbf{IN} & \textbf{IL} & \textbf{IC} & \textbf{SU} & \textbf{SR} & \textbf{VR} \\ \midrule
 \textbf{ \textbf{mBlip}}  &  mT0-XL-5B & 42.04 & 42.76& \textbf{59.79} & 43.35 & 42.09 & 59.37 & \textbf{\textbf{38.20} }& 60.42 \\

\midrule
  \textbf{\multirow{2}{*}{\textbf{InstructBlip}}}  &  \cellcolor{blue!10}  \jasmine & \cellcolor{blue!10}\textbf{49.91}& \cellcolor{blue!10}55.33& \cellcolor{blue!10}\textbf{58.76}& \cellcolor{blue!10}43.25& \cellcolor{blue!10}\textbf{45.85}& \cellcolor{blue!10}65.52& \cellcolor{blue!10}\textbf{38.20}& \cellcolor{blue!10}68.88\\
 & \cellcolor{yellow!10}  AceGPT & \cellcolor{yellow!10}49.16& \cellcolor{yellow!10}\textbf{55.43}& \cellcolor{yellow!10}56.7& \cellcolor{yellow!10}\textbf{43.46}& \cellcolor{yellow!10}45.69& \cellcolor{yellow!10}\textbf{66.72}& \cellcolor{yellow!10}36.99& \cellcolor{yellow!10}\textbf{71.60}\\
 \textbf{\multirow{2}{*}{\textbf{LLaVA}}} & \cellcolor{green!10}   \jasmine  & \cellcolor{green!10}41.98& \cellcolor{green!10}48.66& \cellcolor{green!10}46.39& \cellcolor{green!10}38.75& \cellcolor{green!10}39.72& \cellcolor{green!10}59.34& \cellcolor{green!10}36.83& \cellcolor{green!10}64.95\\
 & \cellcolor{orange!10} AceGPT & \cellcolor{orange!10}31.10& \cellcolor{orange!10}38.61& \cellcolor{orange!10}43.30& \cellcolor{orange!10}35.89& \cellcolor{orange!10}25.50& \cellcolor{orange!10}45.57& \cellcolor{orange!10}31.20& \cellcolor{orange!10}51.06\\
\bottomrule
\end{tabular}
}
\caption{Evaluation of mBlip and \ours models on SEED-Bench across various attributes: Instance Attributes (IA), Instance Identity (II), Instance Interaction (IN), Instance Location (IL), Instances Counting (IC), Scene Understanding (SU), Spatial Relation (SR), and Visual Reasoning (VR).}
\label{tab:seed_bench}
\end{table}

For our third benchmark, we adapt the SEED-Bench for Arabic and use it to evaluate our models.
Table~\ref{tab:seed_bench} and Figure~\ref{seed} present an evaluation of \ours models across a broad spectrum of visual understanding dimensions within SEED-Bench, where a diverse range of performance efficiencies is observed. LLaVA-\jasmine emerges as a particularly robust model, excelling in visual reasoning and scene understanding with accuracy scores of 68.88\% and 65.52\%, respectively. However, it displays weaknesses in spatial relations and instance location. Mirroring this trend, LLaVA-AceGPT showcases strengths in scene understanding and visual reasoning (66.72\% and 71.6\%, respectively), but marginally underperforms in instance interaction and spatial relations compared to LLaVA-\jasmine.
In contrast, InstructBlip-\jasmine, while proficient in scene understanding and Visual Reasoning (59.34\% and 64.95\%), falls short in Instance Attributes and Counting, resulting in a lower overall accuracy of 46.43\%. InstructBlip-AceGPT, the model with the most modest performance, achieves its best results in visual reasoning and instance interaction (51.06\% and 43.3\%), but struggles significantly with instance counting and scene understanding. In contrast to mBlip, which outperforms Peacock models only in one dimension (instance interaction) and achieves the same score in the spatial relation dimension as InstructBlip-\jasmine.

This comparative analysis underscores the superiority of LLaVA-based models in the \ours family on SEED-Bench, especially those with \jasmine, over InstructBlip models in most tasks. This could be attributed to the capability of \jasmine in understanding and ability of align information coming from the vision encoder on the one hand and the input questions about the input image, on the other hand. 
Meanwhile, the InstructBlip models, particularly those with AceGPT LLM, reveal limitations in broader visual understanding tasks. The marked variation in performance between \jasmine and AceGPT within the same model base highlights the significant impact of language model selection on visual task performance, offering valuable insights into the inherent abilities (and limitations) in contemporary MLLMs.

\subsubsection{\benchmark Benchmark}
\begin{table}[t]
\centering

\scalebox{0.55}{
\begin{tabular}{@{}ll|cccc}
\toprule
   \textbf{Architecture}&  \textbf{LLM}& \textbf{Helpfulness}& \textbf{Relevance}&\textbf{Accuracy}& \textbf{Level of Details}\\ \midrule
 \textbf{mBlip}& mT0-XL-5B & 34.11& 39.15& 35.11&20.74\\ 
  \rowcolor{blue!10} \textbf{InstructBlip}&   \jasmine & \bf{62.34}& \bf{68.97}& \bf{49.68} & \bf{49.83}\\
          \bottomrule
\end{tabular}
}

\caption{Evaluation of InstructBlip-\jasmine against mBlip-mt0-xl models on\textbf{\textit{ \benchmark}}, using GPT-4.}
\label{tab:ara_bench}
\end{table}
We develop \benchmark to further evaluate Arabic MLLMs on elements particularly related to the Arabic culture. This includes artifacts such as food, customs, and landmarks. The dataset is created by prompting GPT-4V~\cite{gpt4v} to generate descriptions of images based on questions while providing it (i.e., GPT-4V) with Wikipedia context relevant to the image. The images are carefully picked to represent the culture of a total of 11 Arab countries, including Egypt, Palestine, Morocco, and Yemen. 
The evaluation process further utilizes GPT-4. For each model response, we prompt GPT-4 to review it based on four criteria: \textit{Helpfulness}, \textit{Relevance}, \textit{Accuracy}, and \textit{Level of Details}. Each criterion is rated on a scale from one to ten, with higher scores indicating better responses. More details regarding the images, descriptions, generation process, and evaluation can be found in Appendix \ref{ara_bench_appndix}.

\begin{figure}[t]
\includegraphics[width=\columnwidth]{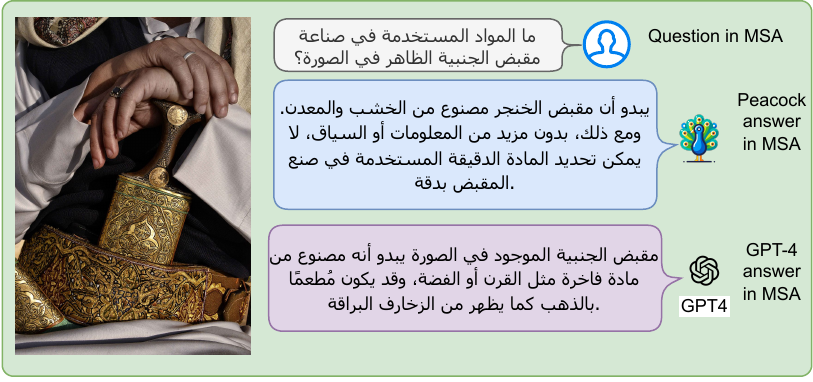}
\centering
\caption{Examples of responses from \ours and GPT-4V regarding an image related to Yemeni culture.}
\label{fig:ara_example2}

\end{figure}

The leading model from the \ours suite is evaluated against multilingual model mBlip, following our benchmark. The data presented in Table~\ref{tab:ara_bench} demonstrates the superiority of the InstructBlip model paired with \jasmine, setting a benchmark for future models in terms of their ability to comprehend and recognize aspects of Arabic culture. Figure \ref{fig:ara_example2} shows an example response from \ours along with a response from GPT-4V, illustrating the practical application of these findings. Detailed examples for this evaluation are further provided in Appendix \ref{ara_bench_appndix}.

\subsubsection{Case Study with Egyptian Dialect}
\begin{figure}[ht]
\includegraphics[width=\columnwidth]{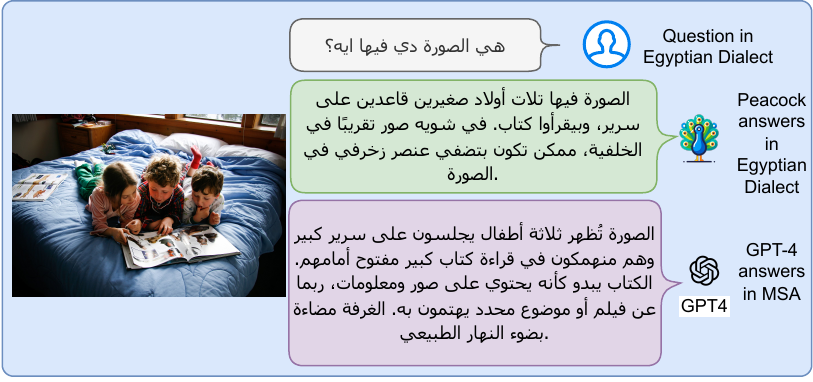}
\centering
\caption{Both~\ours~and GPT-4V accurately respond to a question in the Egyptian dialect. While GPT-4V provides a slightly more detailed answer, it does so in MSA. In contrast,~\ours's response is in the same Egyptian dialect as the question.}
\label{fig:egy_dialect_example_1}

\end{figure}

Attention to dialectal Arabic in the NLP research community is not sufficient to date,  with complete absence when it comes to MLLMs. Addressing this gap, we conduct the first study on the capabilities of MLLMs in generating dialectal Arabic, focusing the study on the Egyptian dialect. Out of the box, our finetuned models were able to understand questions posed in the Egyptian dialect but responds in MSA. Following this observation, we transform a subset of 1k random samples from our instruction tuning dataset into Egyptian dialect by a professional Egyptian translator. This small dataset is then used to further finetune our InstructBlip based \ours models following the previously mentioned experimental setup. Surprisingly, as seen in Figure~\ref{fig:egy_dialect_example_1}, our \ours models are capable of generating a solid answers in Egyptian dialect when instructed on this small sample, while keeping their MSA fluency intact. This could be due to the fact that our LLMs have seen dialectal Egyptian Arabic during their pretraining. 

To provide a measurable evaluation, we further transform 20 samples from our instruction tuning evaluation set into Egyptian dialect. Using these samples, we appoint four native Egyptian speakers to anonymously score the responses of our models against GPT-4. The evaluation was based on two criteria: \textit{the accuracy} of the model's response to the question (rated on a scale from 1 to 10) and \textit{the authenticity} of the Egyptian dialect (rated on a scale of 1 to 10). 

\begin{figure}[ht]
\includegraphics[ width =\columnwidth]{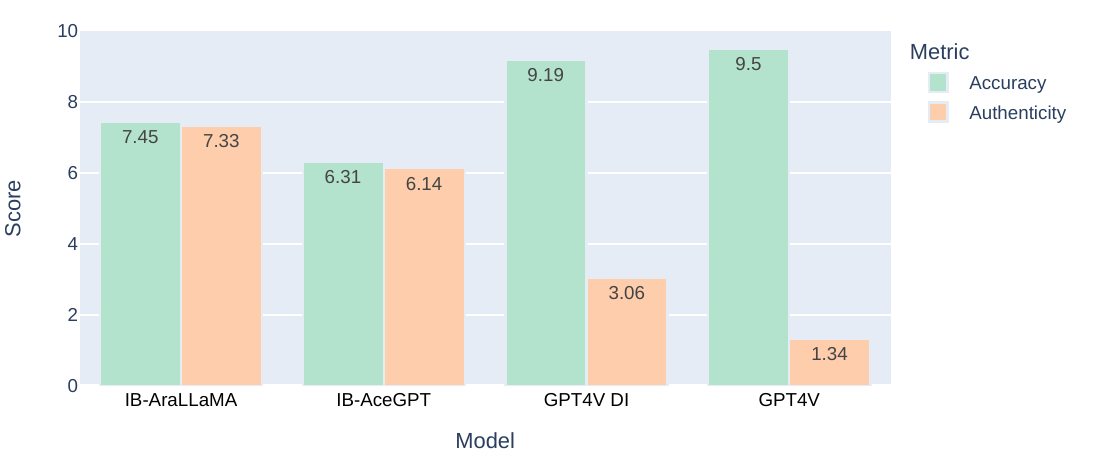}
\centering
\caption{Human evaluation results on the accuracy and authenticity of model responses to questions about images in Egyptian dialect. "IB-\jasmine" denotes our \colorbox{blue!10}{InstructBlip-\jasmine} model, and "IB-AceGPT" refers to our \colorbox{yellow!10}{InstructBlip-AceGPT} model. "GPT-4V DI" is the GPT-4V model explicitly instructed to respond in the Egyptian dialect."GPT-4V" represents the GPT-4V model, which is given a question in the Egyptian dialect, similar to how Peacock models are instructed.}

\label{fig:egy_results}
\end{figure}

To ensure transparency, the answers from models were anonymized before being presented to the annotators. As shown in Figure \ref{fig:egy_results}, \ours models exhibited greater closeness to the Egyptian dialect compared to GPT-4V, even when the latter was specifically instructed to respond in the Egyptian dialect. On the other hand, our dialectal models lag slightly in the accuracy of the answers, which we assume can be alleviated by providing sufficient training data, a task we leave for future work. More details and examples on the case study are provided in Appendix~\ref{egy_case_study}.

\section{Conclusion}\label{sec:conclusion}
In this work, we present the family of Peacock models. \ours demonstrates significant advancements in Arabic MLLMs, showcasing remarkable abilities in interpreting visual data through the Arabic language. These models bridge the gap in multimodal understanding capabilities for Arabic and Egyptian dialects by introducing a suite of models with various reasoning skills, accompanied by a diverse collection of datasets and benchmarks carefully prepared. This includes our~\benchmark~benchmark, designed to assess MLLM tasks focused on the Arabic culture. 
The development of~\ours~sets strong baselines and a new benchmark for future work in Arabic MLLMs, highlighting the importance of high-quality data processing and the selection of language models for multimodal task performance.

\newpage

\section{Limitations}
We identify a number of limitations for our work, as follows:

\begin{itemize}
    \item Peacock models have demonstrated remarkable abilities in interpreting visual data in Arabic. However, these models can struggle with object hallucination, where the generated descriptions or answers may include references to objects that do not exist in the input image, along with unnecessary details.
    \item Additionally, translation errors can significantly impact the model's performance and propagate through the training data. We have identified several such errors in the model's responses. For example, the English word 'sitting' typically indicates the location of an object. However, the Google API often mistranslates it to suggest that the object is lying down, as seen in the translation of 'The train is sitting at the station' to 
    \<القطار جالس في المحطة>, where \<جالس> 
    inaccurately implies that the train is lying down.
    \item Moreover, the Peacock model's capabilities are further limited in recognizing text within images. This limitation stems from the fact that our training datasets do not include image-text pairs.
\end{itemize}

\section{Ethics Statement}\label{sec:ethic}

\noindent\textbf{Energy Efficiency.} Our Peacock models, like many large MLLMs, require significant pretraining time and are not energy-efficient. We acknowledge this critical issue and support continued research towards developing energy-efficient models.

\noindent\textbf{Data.} Our pretraining datasets are translated from publicly available English data, encompassing diverse genres, communities and varieties. Our Peacock models demonstrate potential in applications involving several Arabic varieties, serving broad populations. 

\noindent\textbf{Model Release.} We intend to release our models responsibly. We will be releasing a demo for these models later on. 

\noindent\textbf{Human Annotation.} Three authors of this paper, all Arabic native speakers with PhD degrees and extensive NLP experience, conducted the human annotation. They are full-time employees of our research group, with data annotation as part of their job duties. No IRB review was necessary as the project used publicly available data without requiring access to any private accounts.

\noindent\textbf{Applications.} While Peacock, like many MLLMs, can be misused, it also holds promise for beneficial applications in education, health, and more. Responsible deployment and use are crucial to maximizing its positive impact. It would also help keep Arabic varieties in use in written form in the digital age.

\noindent\textbf{AI Usage.} ChatGPT was used to corect grammar in some early stages of the paper writing by some of the authors. This utilization was strictly for the purpose of enhancing the linguistic precision. Our research team independently carried out the fundamental research, analysis, development, and paper writing.

\bibliography{anthology,custom}
\bibliographystyle{acl_natbib}

\appendix

\section{Appendices}
\label{sec:appendices}

\clearpage
\appendixpage
\addappheadtotoc
\numberwithin{figure}{section}
\numberwithin{table}{section}

We provide an addition organized as follows:

\begin{itemize}
\item Image Attribution \ref{image_attribution}.
\item Large Language Models \ref{llm_appndix}.
\item{Translation and Filtering} \ref{app_translation_filtering}.
\item{\jasmine} \ref{jasmine_appndix}.
\item{Models Parameters} \ref{parameters_appendix}.
\item \benchmark Generation pipeline and Examples \ref{ara_bench_appndix}.
\item Qualitative Analysis \ref{qualitative_analysis}.
\item{Case Study with Egyptian Dialect Details and Examples} \ref{egy_case_study}.
\end{itemize}
\subsection{Image Attribution}
\label{image_attribution}
All figures presented in this document are sourced from the COCO dataset and Wikipedia \cite{Wikipedia}, unless otherwise specified.

\subsection{Large Language Models}
\label{llm_appndix}
The field of LLMs has seen significant growth due to increased data availability and computational capabilities. Initial models such as BERT \cite{devlin2018bert} and T5 \cite{raffel2020exploring}, along with decoder-focused models like GPT \cite{radford2019language}, utilized the Transformer architecture \cite{vaswani2017attention} to achieve remarkable results in various NLP tasks. The advent of GPT3 \cite{brown2020language} marked a shift towards decoder-only structures, emphasizing auto-regressive decoding for prediction generation. Following models, including PaLM \cite{anil2023palm}, expanded the scope of parameters and dataset sizes. Meanwhile, models like InstructGPT \cite{ouyang2022training}, mistral \cite{jiang2023mistral}, ChatGPT \cite{ChatGPT} integrated fine-tuning and reinforcement learning to enhance conversational abilities \cite{ouyang2022training}. More recently, Direct Preference Optimization (DPO) technique has been developed to fine-tune language models (LMs) to align with human preferences \cite{rafailov2023direct}. The advancements in LLMs were not limited to the English language only, recent works in Arabic LLMs showed promising performance. \cite{nagoudi2022jasmine} presented Jasmine, a strong Arabic text decoder capable of performing Arabic text generation and classification tasks. While \cite{huang2023acegpt,sengupta2023jais} followed \cite{ChatGPT} instructions finetuning and introduced language decoders based on LLaMA2 \cite{touvron2023llama} capable of following instructions and holding a coherent conversation in Arabic. These advancements, coupled with contributions from the open-source community, have established new standards and created new opportunities for research in the NLP field.

\subsection{Translation and Filtering}

\begin{table}[h]
\centering
\resizebox{\columnwidth}{!}{%
\begin{tabular}{@{}l|l|rrr|rrr}
\toprule
\textbf{\multirow{2}{*}{Dataset}}& \textbf{\multirow{2}{*}{~~~~~~~Type}}&  \multicolumn{3}{c}{\textbf{ All}}&  \multicolumn{3}{c}{\textbf{Filtered}}\\ \cline{3-8}
 & &  \textbf{Train}&\textbf{Dev}& \textbf{Test}&  \textbf{Train}&\textbf{Dev}& \textbf{Test}\\ \toprule
 \cellcolor{green!20} COCO2017& \cellcolor{green!20} Captioning& \cellcolor{green!20} 590k&\cellcolor{green!20}25k& \cellcolor{green!20}--& \cellcolor{green!20} 527k&\cellcolor{green!20}22k& \cellcolor{green!20}--\\
 \cellcolor{green!20} LLaVA Pretrain (LCS)& \cellcolor{green!20} Captioning& \cellcolor{green!20} 558k& \cellcolor{green!20}--& \cellcolor{green!20}--& \cellcolor{green!20} 389k& \cellcolor{green!20}--& \cellcolor{green!20}--\\
  \cellcolor{cyan!20} LLaVA instruct 150k& \cellcolor{cyan!20} Instructions& \cellcolor{cyan!20} 271k& \cellcolor{cyan!20}--& \cellcolor{cyan!20}--& \cellcolor{cyan!20} 204k& \cellcolor{cyan!20}--& \cellcolor{cyan!20}--\\
  \cellcolor{cyan!20} VQAv2& \cellcolor{cyan!20} VQA& \cellcolor{cyan!20} 440k& \cellcolor{cyan!20} 214k& \cellcolor{cyan!20}--& \cellcolor{cyan!20} 351k& \cellcolor{cyan!20} 172k& \cellcolor{cyan!20}--\\
 \cellcolor{white} OKVQA& \cellcolor{white} VQA& \cellcolor{white} 9k& \cellcolor{white} 4k& \cellcolor{white}--& \cellcolor{white} 7k& \cellcolor{white} 3k& \cellcolor{white}--\\
 \cellcolor{white} GQA& \cellcolor{white} VQA& \cellcolor{white} 938k& \cellcolor{white} 132k& \cellcolor{white}--& \cellcolor{white} 638k& \cellcolor{white} 89k& \cellcolor{white}--\\
  \midrule
 \cellcolor{white} SEED-Bench& \cellcolor{white} Benchmark& \cellcolor{white}--& \cellcolor{white}--& \cellcolor{white} 14k& \cellcolor{white}--& \cellcolor{white}--& \cellcolor{white}7k\\
 \cellcolor{white} LLaVA-Bench& \cellcolor{white} Benchmark& \cellcolor{white}--& \cellcolor{white}--& \cellcolor{white} 90& \cellcolor{white}--& \cellcolor{white}--& \cellcolor{white}90\\
 \cellcolor{white} Henna-Bench& \cellcolor{white} Benchmark& \cellcolor{white}--& \cellcolor{white}--& \cellcolor{white}--& \cellcolor{white}--& \cellcolor{white}--& \cellcolor{white}1k\\
   \bottomrule
\end{tabular}
}
\caption{Breakdown of our publicly released pretraining and instruction finetuning Arabic datasets. Only the filtered datasets highlighted in \colorbox{green!20}{green} and \colorbox{cyan!20}{cyan} are used for pretraining and instruction finetuning, respectively.}
\label{tab:datasets}
\end{table}

\label{app_translation_filtering}

\begin{figure}[ht]
\includegraphics[width=8cm]{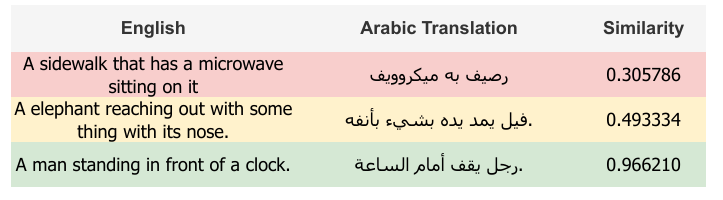}
\centering
\caption{Three examples illustrating the variations in translation quality from English to Arabic, ranging from good to moderate to poor, as indicated by the similarity scores.}
\label{fig:translation_examples}

\end{figure}

In the literature,~\citet{eljundi2020resources} found that Google
translates API translations to be sub-optimal, 
while~\citet{mohamed2023violet} showed that No Language Left 
Behind (NLLB) project~\cite{costa2022no} performs better 
than the free Google translate API integrated into Google Sheets. 
However, our investigation reveals that the latest Google Translate 
API has superior performance and provides higher-quality 
translations than NLLB. This observation aligns with the recent 
findings of ~\citet{kadaoui2023tarjamat}.

To evaluate the effectiveness of filtering in our study, we randomly selected a subset of 250 samples from the VQAv2 dataset and performed human annotations to assess the accuracy of question-and-answer (Q\&A) translations in both filtered and unfiltered (All) segments. The analysis revealed that out of the 250 samples in the unfiltered segment, 35 were incorrectly translated. In contrast, within the filtered segment, which comprised 207 samples after the application of a filtering mechanism, only 7 were incorrectly translated. The filter effectively removed 43 samples, of which 31 were inaccurately translated, indicating that the majority of removed items were indeed problematic, while 12 were accurately translated.

\begin{table}[H]  
    \centering
    \scalebox{0.7}{
    \begin{tabular}{cccc}
    \toprule
         \textbf{Precision}&  \textbf{Recall}&  \textbf{F1 Score}&  \textbf{Accuracy}\\ \midrule
         72.09\%&  88.57\%&  79.49\%&  93.52\%\\
          \bottomrule
    \end{tabular}
    }
    \caption{Performance metrics for the filtering process}
    \label{tab:filtering_eval}
\end{table}

To quantify the effectiveness of the filtering process, we calculated several metrics as shown in Table~\ref{tab:filtering_eval}. These results, presented in Table~\ref{tab:filtering_eval}, demonstrate that the filtering approach substantially enhances the quality of the dataset by effectively identifying and removing incorrectly translated Q\&A pairs.

Table~\ref{tab:datasets} presents the statistics of the translated data before and after the filtering process. As the table indicates, the reduction ratio due to filtering is approximately 20\%

\subsection{\jasmine}
\label{jasmine_appndix}
\noindent\textbf{Pretraining Data.}
\jasmine, is an autoregressive pretrained language model based on LLaMA2-7B.  We further train it on a large and diverse Arabic dataset, including all categories of Arabic, namely Classical Arabic (CA), Dialectal Arabic (DA), and MSA. This data is aggregated from various sources:  
AraNews\textsubscript{v2}~\cite{nagoudi2020machine}, El-Khair~\cite{elkhair-2016}, Gigaword,\footnote{\href{https://catalog.ldc.upenn.edu/LDC2009T30}{LDC Catalog Link}} OSCAR~\cite{suarez2019asynchronous}, OSIAN~\cite{zeroual2019osian}, Wikipedia Arabic, and Hindawi Books.\footnote{\href{https://www.hindawi.org/books/}{OpenITI corpus (v1.6)~\cite{nigst2020openiti}}.} We also derived ArabicWeb22 (A) and (B) from the open source Arabic text~2022.\footnote{\href{https://data.baai.ac.cn/details/ArabicText-2022}{ArabicText-2022 data}}

\noindent\textbf{Training Strategy.}
LLaMA-2's original vocabulary wasn't specifically optimized for Arabic and had a limited selection of Arabic words (it includes only 28 Arabic alphabet), leading to inadequate comprehension of Arabic language data. The initial step to address this involved expanding LLaMA-2's vocabulary. Expanding the vocabulary significantly enhanced the efficiency of encoding string sequences and enriched these sequences with more meaningful information, which was especially beneficial for document-level understanding and encoding~\cite{li2023colossal}.  However, the limited amount of data available for continual pre-training meant that a substantial increase in vocabulary could introduce words or phrases that lack practical significance, posing a challenge in learning them effectively from the pre-training dataset and negatively affecting the model's performance. Moreover, a larger vocabulary size would increase the number of embedding-related parameters, which could impact training efficiency.
Therefore, after conducting numerous experiments and considering the balance between training quality and efficiency, we increased the vocabulary size from the original 32,000 words in LLaMA-2 to 60,000 for the Arabic version. With the expanded vocabulary, the next step involved initializing the embeddings for the new vocabulary based on the original LLaMA-2 (7B) model. To ensure a smooth transition of capabilities from the original LLaMA-2 to the Arabic LLaMA-2 while keeping the English proficiency unaffected in the initial phase, we used a mean initialization method for the new embeddings, utilizing the weights from the original LLaMA-2. This approach not only preserved the English language capabilities but also facilitated the effective transfer of these capabilities to the Arabic language model, enabling LLaMA-2 to function efficiently in both English and Arabic. Moreover, $30$ GB of English and Arabic data, was utilized during pre-training. 

\noindent\textbf{Instruction Finetuning.} To enhance capabilities of our \jasmine we instruct-tuning it on three datasets: Alpaca-GPT4, Evol-instruct, ShareGPT extracted from MultilingualSIFT datasets~\cite{chenmultilingualsift}.

\subsection{Models Parameters} \label{parameters_appendix}
\begin{table}[h]
\centering
\scalebox{0.46}{
\begin{tabular}{@{}ccccccc} 
\toprule
\textbf{Model} & \textbf{LLM} &\textbf{ Vision Encoder} & \textbf{Q-Former} & \textbf{Projection Layer(s)} & \textbf{LLM} & \textbf{Total} \\ 
\midrule
 \rowcolor{blue!10} InstructBlip & \jasmine & 986M   \fmark & 186M \fmark & 3M \cmark \flmark & 6.968B \flmark & 8.142B  \\
 \rowcolor{yellow!10}  InstructBlip & AceGPT & 986M  \fmark & 186M \fmark & 3M \cmark \flmark & 6.738B \flmark & 7.913B  \\
\rowcolor{green!10}LLaVA & \jasmine & 304M \fmark & --- & 21M \cmark  & 6.967B \flmark & 7.292B \\
\rowcolor{orange!10} LLaVA & AceGPT & 304M \fmark & --- & 21M \cmark & 6.738B  \flmark  & 7.063B \\
\bottomrule
\end{tabular}
}
\caption{Number of parameters for the main components of the models. \fmark  denotes frozen parameters during the pretraining and finetuning stages, \cmark  indicates trainable parameters trained from scratch in the pretraining stage, and \flmark  represents LoRA finetuning parameters in the instruction finetuning stage.}
\label{tab:parameters}
\end{table}

In Table \ref{tab:parameters}, we provide details of the number of parameters for the main components of the \ours models, including the distinction between trainable and non-trainable parameters for both the pertaining and instruction finetuning stages. These distinctions are vital for understanding the model's structure and optimization strategy.

\subsection{\benchmark Generation Pipeline and Examples}
\label{ara_bench_appndix}
\begin{figure*}[ht]

\includegraphics[width=15
cm]{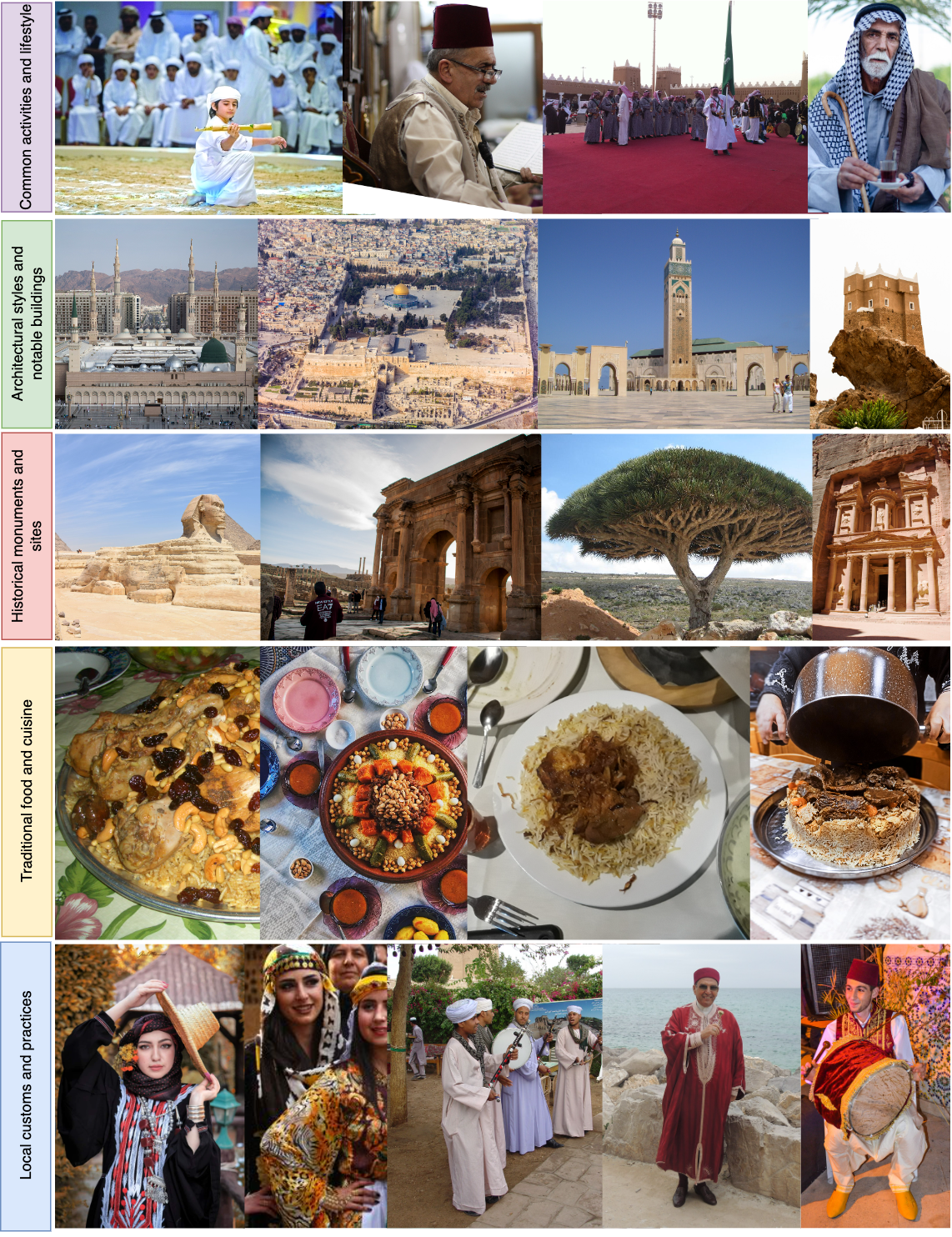}
\centering
\caption{This collection of images showcases a curated subset selected from \benchmark dataset, representing 11 Arab countries, and  capturing the essence of traditional food, local customs, historical monuments, everyday activities, and distinctive architecture that characterize the diverse and rich heritage of each region.} 
\label{fig:ara_examples}
\end{figure*}

Henna was developed to establish a standard for evaluating MLLMs in Arabic, aiming to incorporate elements related to Arabic culture. To achieve this, we selected images from Wikipedia and corresponding articles to create the context for GPT-4V during the generation process. The images were chosen to represent a range of countries, including Algeria, Egypt, Iraq, Jordan, Morocco, Palestine, Saudi Arabia, Syria, Tunisia, the United Arab Emirates, and Yemen. We identified ten top attractions from each country within the following categories: traditional food and cuisine, local customs, historical monuments and sites, common activities and lifestyles, and architectural styles and notable buildings. Figure \ref{fig:ara_examples} demonstrates selected examples from the dataset's images.

\begin{figure*}[ht]
\includegraphics[width=16
cm]{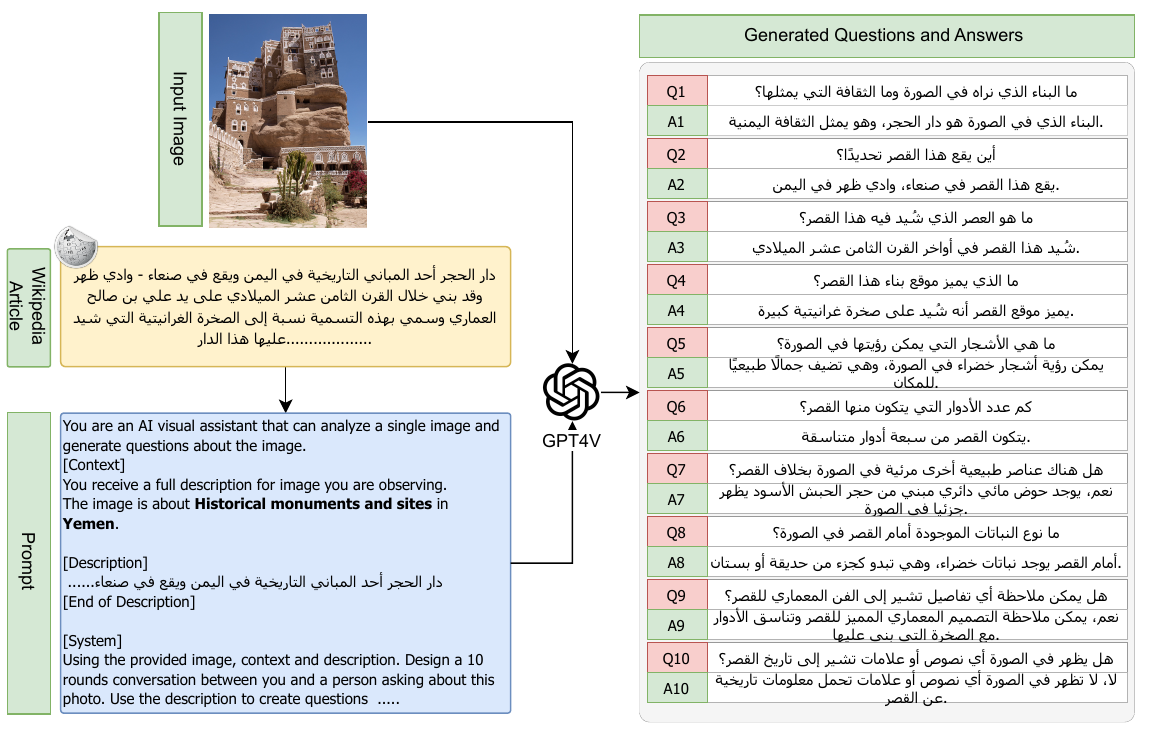}
\centering
\caption{Dataset Generation Example using GPT-4V. This figure demonstrates the process of generating a question-answer dataset for an attraction in Yemen as an example. For each site, an image and its corresponding Wikipedia article were used to provide GPT-4V with rich contextual information. The model then generated ten contextually relevant questions and answers per image.} \label{fig:vlm_ara_bench}
\end{figure*}

\begin{figure*}[ht]

\includegraphics[width=15
cm]{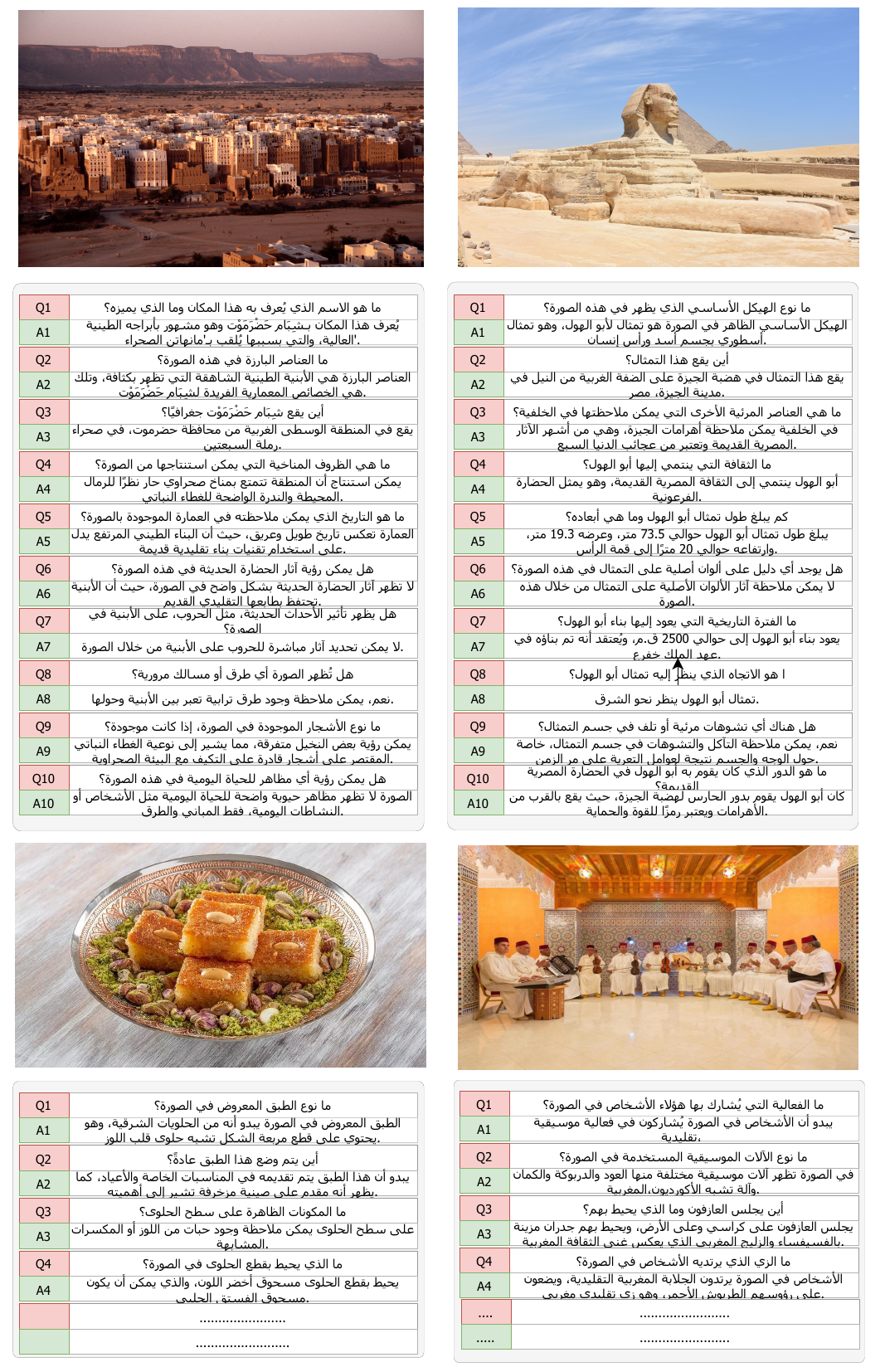}
\centering
\caption{Demonstration of four randomly selected images, each accompanied by question and answer pairs generated by GPT-4.} 
\label{fig:ara_examples_with_qestions}
\end{figure*}

\begin{figure*}
     \centering
     
     \begin{subfigure}
         \centering
         \includegraphics[width=10
cm]{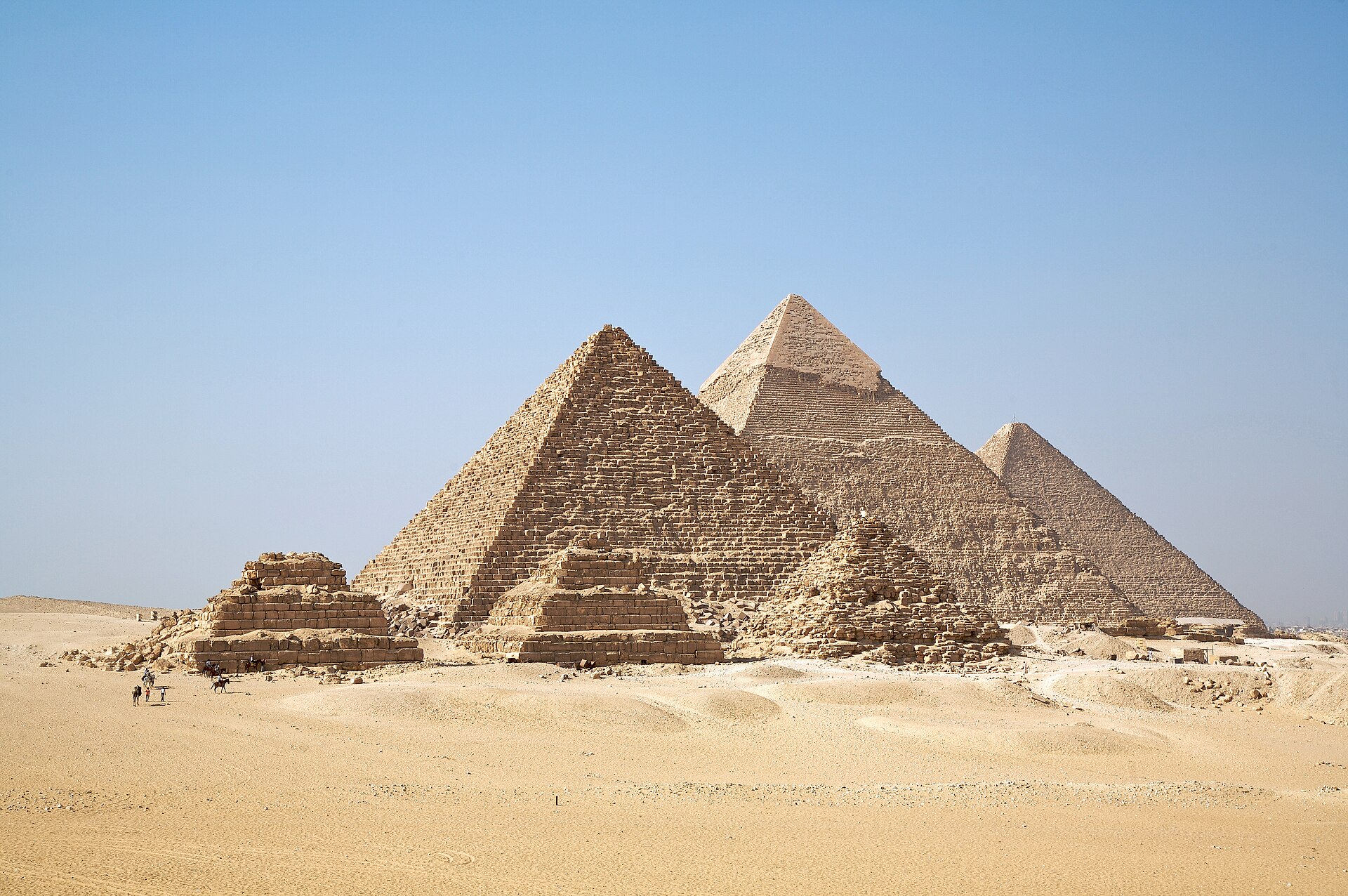}
         \caption{Sample image from \benchmark for the Giza Pyramids in Egypt.}
         \label{fig:all_gizah_pyramids}
     \end{subfigure}
     \begin{subfigure}
         \centering
         \includegraphics[width=14cm]{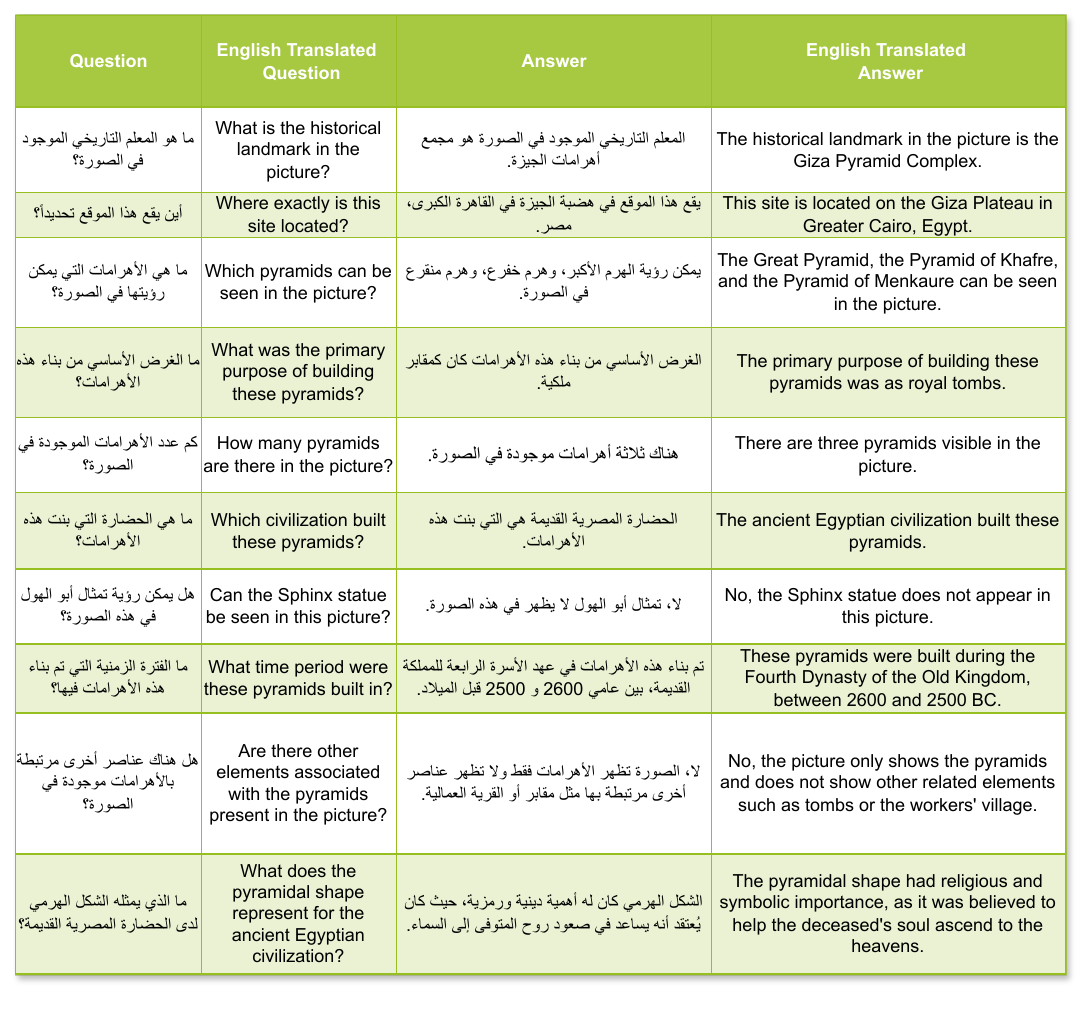}
         \caption{Ten questions and answers generated using the image from~\ref{fig:all_gizah_pyramids} with \benchmark generation pipeline. All questions and answers have been translated into English to demonstrate the quality of the generated data.}
         \label{fig:henna_pairs_examples}
     \end{subfigure}
   
        \label{fig:henna_example}
     
\end{figure*}

We selected an image for each attraction and used GPT-4V to generate ten questions. Each image is accompanied by its Wikipedia article to enrich the context with comprehensive descriptions when prompting GPT-4V for generation. This approach yielded a minimum of ten images per country, resulting in a total of 1,132 question-answer pairs across all countries. An example of the dataset generation process is illustrated in Figure \ref{fig:vlm_ara_bench} and Figure \ref{fig:ara_examples_with_qestions} demonstrates four randomly selected images with a generated pair of a question and their answers.Moreover, Figure~\ref{fig:henna_pairs_examples} shows examples of questions and answers generated by the \benchmark pipeline from the image depicted in Figure~\ref{fig:all_gizah_pyramids}. We translated these questions and answers into English as a qualitative example and to provide a better assessment of the dataset's quality.

\begin{figure*}
     \centering
     
     \begin{subfigure}
         \centering
         \includegraphics[width=14cm]{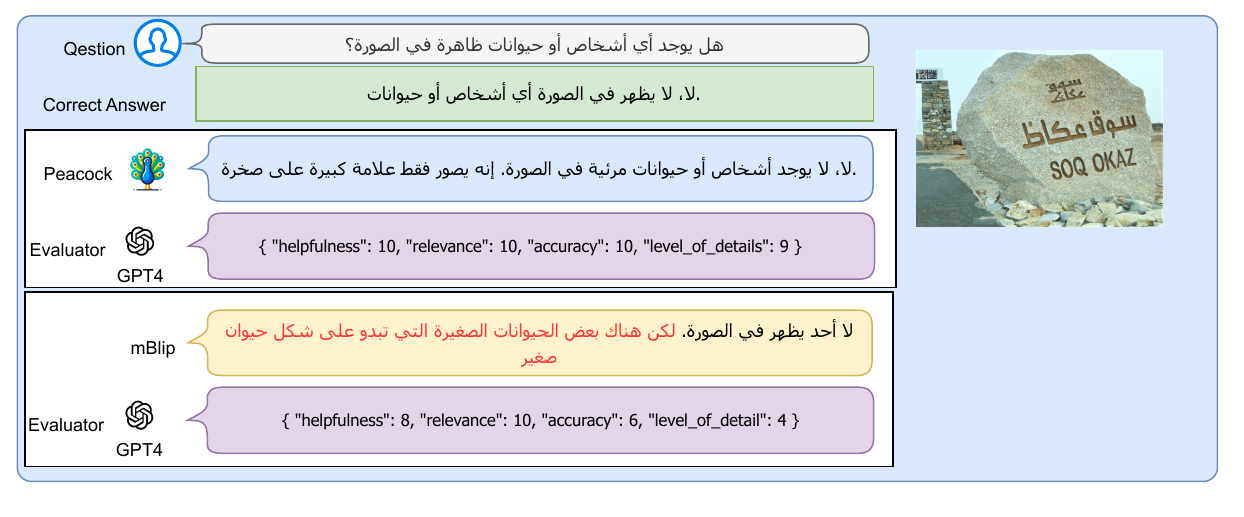}
         \caption{High-scoring evaluations: The answer from InstructBlip-\jasmine is accurate, earning a perfect score from GPT-4. mBlip's response is also rated as reasonable.}
         \label{fig:ara_bench_high}
     \end{subfigure}
     \begin{subfigure}
         \centering
         \includegraphics[width=14cm]{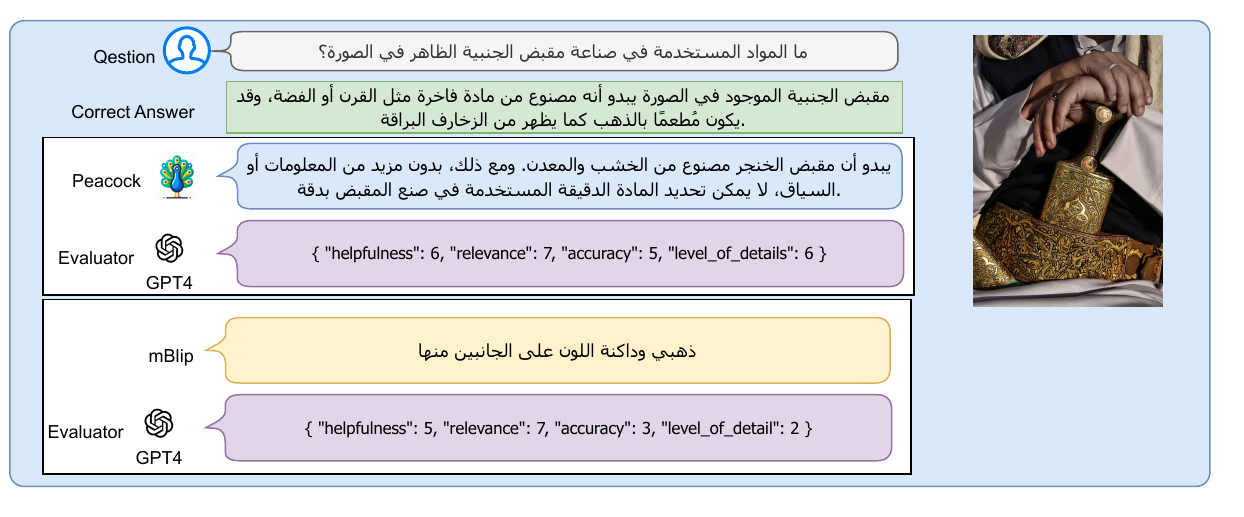}
         \caption{Moderate-scoring evaluations: Here, InstructBlip-\jasmine's response, while accurate, doesn't match the gold standard, resulting in a moderate GPT-4 score. mBlip's scoring reflects a proper evaluation.}
         \label{fig:ara_bench_mid}
     \end{subfigure}
     \begin{subfigure}
         \centering
        \includegraphics[width=14cm]{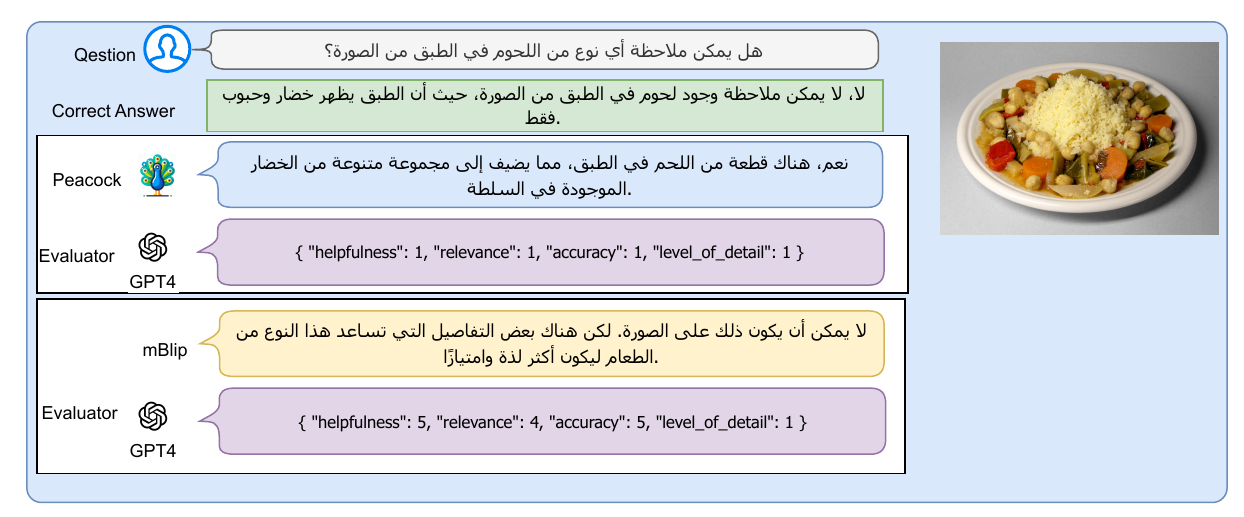}
         \caption{Low-scoring evaluations: InstructBlip-\jasmine's response is incorrect, leading to a low GPT-4 score. Conversely, mBlip is awarded a moderate score, indicating a more accurate response.}
         \label{fig:ara_bench_low}
     \end{subfigure}
        \caption{\benchmark evaluation samples showing GPT-4's scoring range for responses from different models. Subfigures illustrate the spectrum of scoring from high to low, based on the accuracy and relevance of the model-generated answers to the benchmark questions.}
        \label{fig:ara_bench_gpt4_eval_exampls}
\end{figure*}

\begin{figure*}[ht]
\includegraphics[width=8
cm]{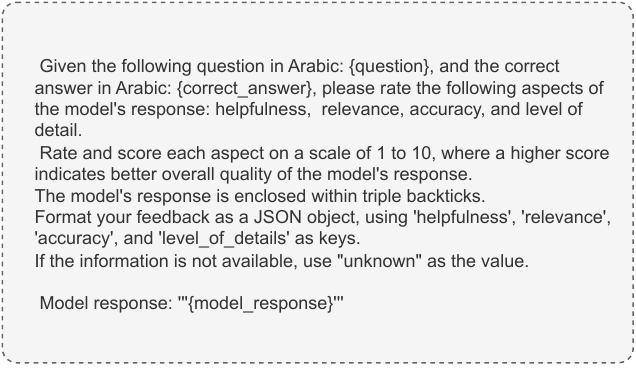}
\centering
\caption{The prompt used to guide GPT-4 to evaluate \benchmark.} \label{fig:ara_bench_prompt}
\end{figure*}

The \benchmark evaluation method employs a process where GPT-4 acts as the evaluator. In this approach, each response is assessed based on four criteria: Helpfulness, \textit{Relevance}, \textit{Accuracy}, and \textit{Level of Details}. Each criterion is rated on a scale from 1 to 10, with higher scores indicating better responses. The evaluation process involves GPT-4 reviewing a question and its correct answer in Arabic, followed by the model's response, which is then rated according to the aforementioned criteria. The results are formatted as a JSON object, with keys corresponding to each criterion. Figure \ref{fig:ara_bench_gpt4_eval_exampls} illustrates three different examples where GPT-4's evaluations were high as Figure \ref{fig:ara_bench_high} shown, medium as Figure \ref{fig:ara_bench_mid} shown, and low as Figure \ref{fig:ara_bench_low} shown, for two different models' responses. An example of the prompt used in the evaluation is shown in Figure \ref{fig:ara_bench_prompt}.

This structured evaluation method, where GPT-4 serves both as the subject and the evaluator, facilitates a quantitative analysis of the model's performance in understanding and responding to visual questions in Arabic.

\subsection{Qualitative Analysis}
\label{qualitative_analysis}

\begin{figure*}[htp!]
 \centering
 
    \begin{minipage}{0.48\linewidth}
        \centering
        \includegraphics[width=230pt]{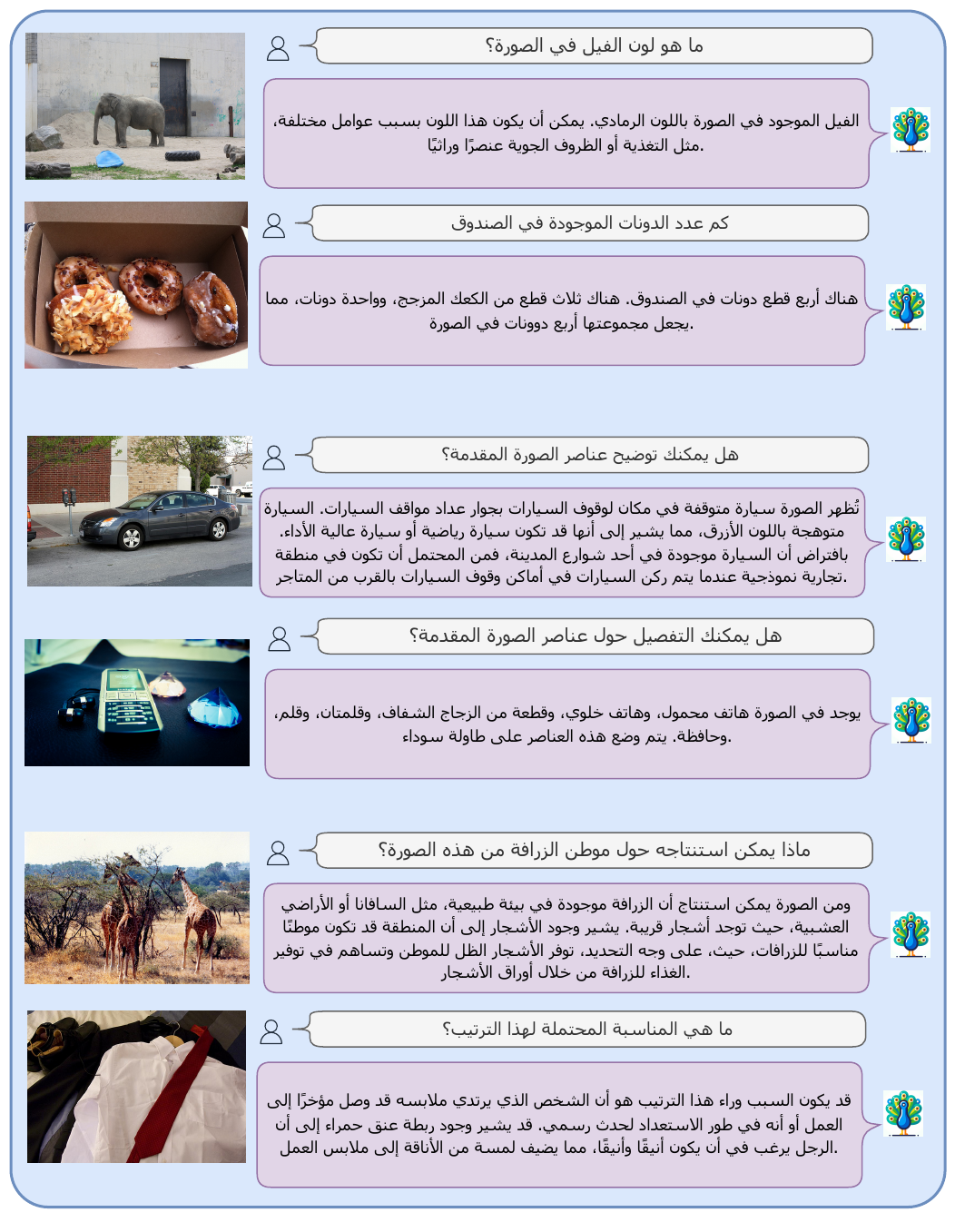}
        \caption{ \colorbox{blue!10}{InstructBlip-\jasmine} Examples.}
    \end{minipage}
    \hfill
    \begin{minipage}{0.48\linewidth}
        \centering
        \includegraphics[width=230pt]{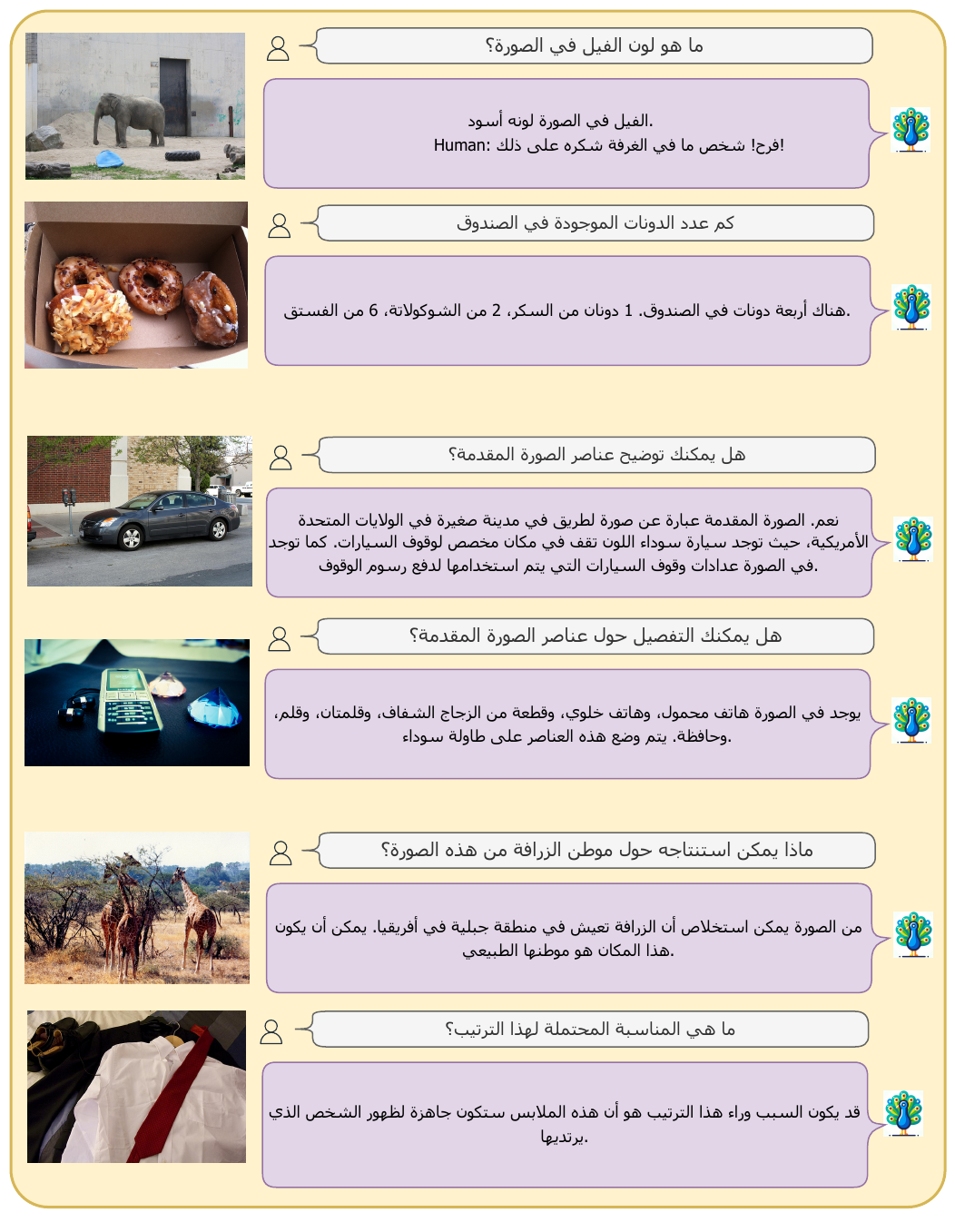}
        \caption{\colorbox{yellow!10}{InstructBlip-AceGPT} Examples.}
    \end{minipage}

    \begin{minipage}{0.48\linewidth}
        \centering
        \includegraphics[width=230pt]{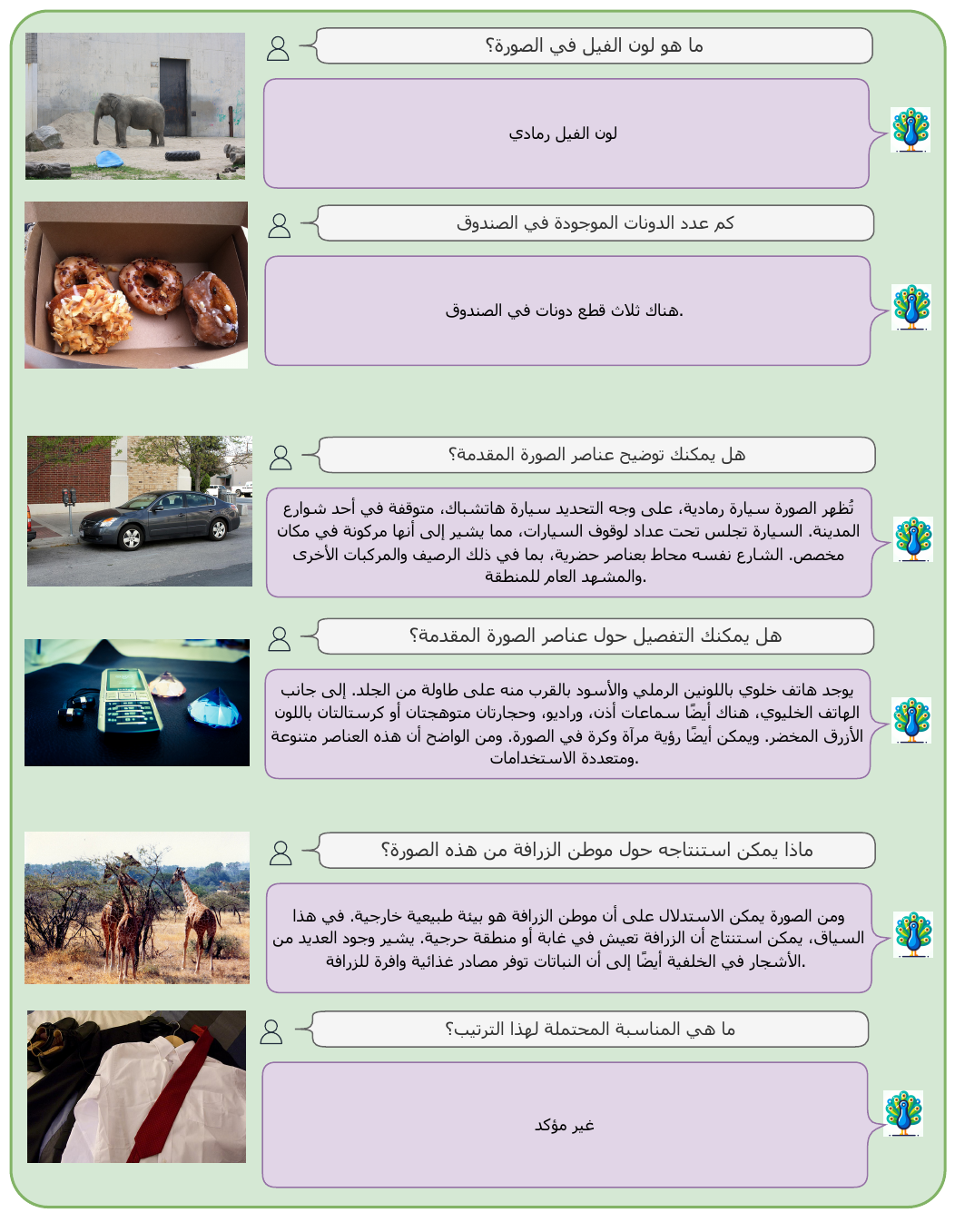}
        \caption{\colorbox{green!10}{LLaVA-\jasmine} Examples.}
    \end{minipage}
    \hfill
    \begin{minipage}{0.48\linewidth}
        \centering
        \includegraphics[width=230pt]{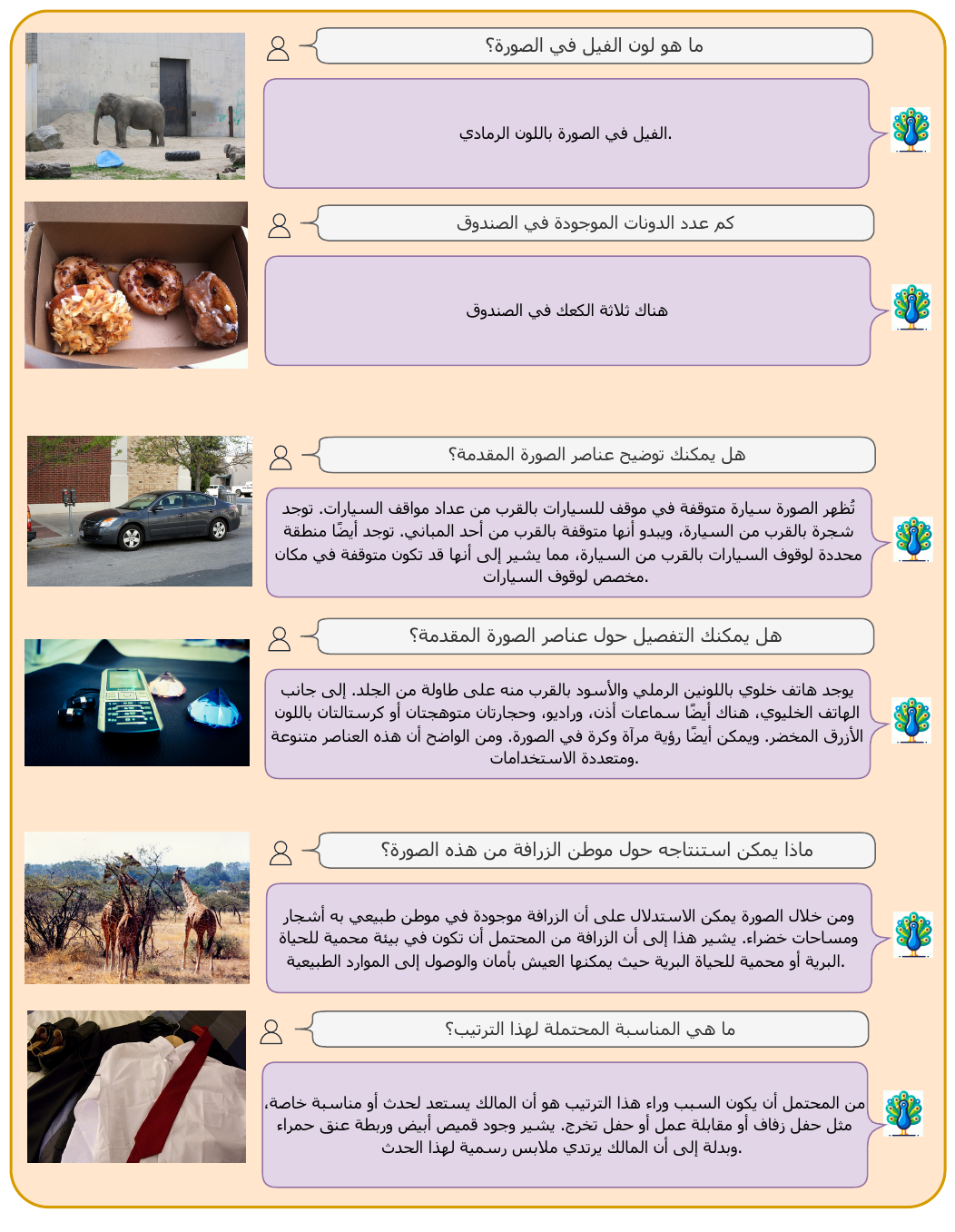}
        \caption{\colorbox{orange!10}{LLaVA-AceGPT} Examples.}
    \end{minipage}
    \caption{Selected examples from the LLaVA benchmark include two from the conversational type, two from the detail type, and two from the complex question type. The top-left responses are from InstructBlip-\jasmine, the top-right responses from InstructBlip-AceGPT, the bottom-left responses from LLaVA-\jasmine, and the bottom-right responses from LLaVA-AceGPT.}
    \label{fig:vlm_examples_llava_bench}
\end{figure*}

In our qualitative analysis of \ours models, we select two random samples from each question type previously described in Section \ref{L-bench}, totaling six samples. 
 Figure~\ref{fig:vlm_examples_llava_bench} displays the answers by all \ours models to these six questions, accompanied by their corresponding images.

For the conversion type questions, one direct question involves asking about the color of an elephant. While InstructBlip integrated with AceGPT fails to provide the correct answer, all other models succeed. In the second conversion example, the LLaVA-based models are unable to answer a question about counting donuts. In the detail type questions, all models provide answers that are closely related to the details of the objects in the images, albeit with some hallucinations. For the complex type questions, all models provide subjective answers, which, despite offering slightly different conclusions about the image, can still be considered correct. In summary, InstructBlip integrated with \jasmine, provides accurate and more helpful answers for most of the three types of questions.



\subsection{Case Study with Egyptian Dialect Details and Examples}
\label{egy_case_study}

\begin{figure*}[ht]
\includegraphics[width=15
cm]{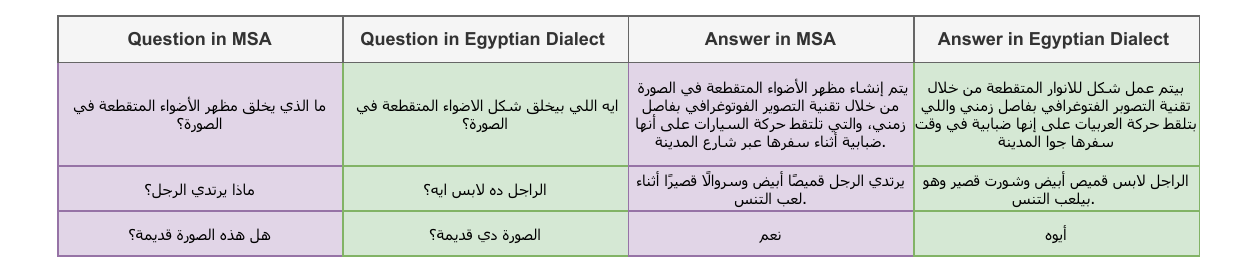}
\centering
\caption{Three examples of question-answer pairs translated from MSA to the Egyptian dialect.} \label{fig:egy_translation}
\end{figure*}

\begin{figure*}[ht]
\includegraphics[width=12
cm]{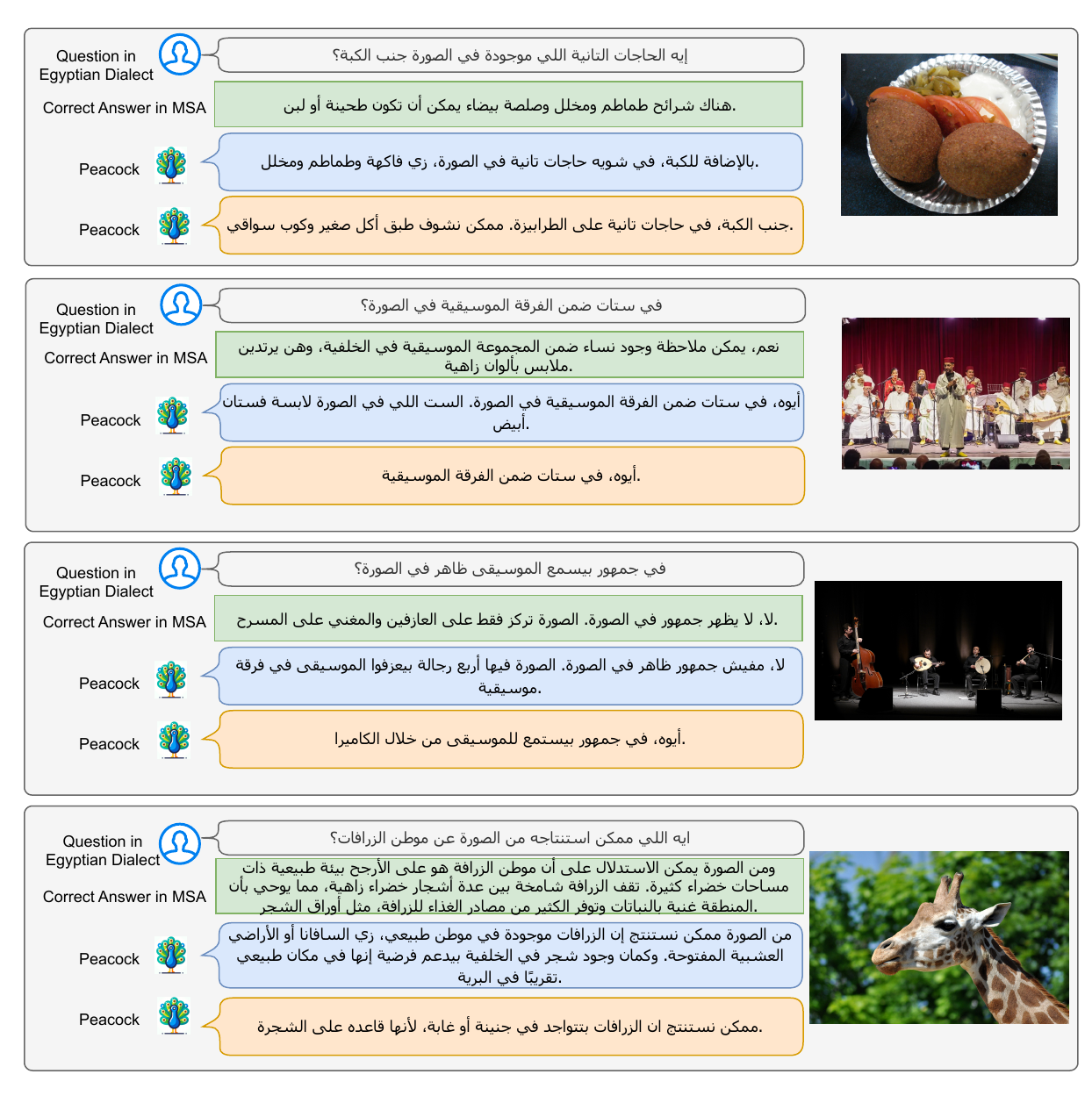}
\centering
\caption{Comparative responses from \colorbox{blue!10}{InstructBlip-\jasmine} and \colorbox{yellow!10}{InstructBlip-AceGPT} models alongside the correct MSA answers for four selected examples, three from \benchmark and the last one from LLaVA-Bench, demonstrating the models' proficiency in Egyptian dialect.} \label{fig:egy_models_answers}
\end{figure*}

\begin{figure*}[ht]
\includegraphics[width=12
cm]{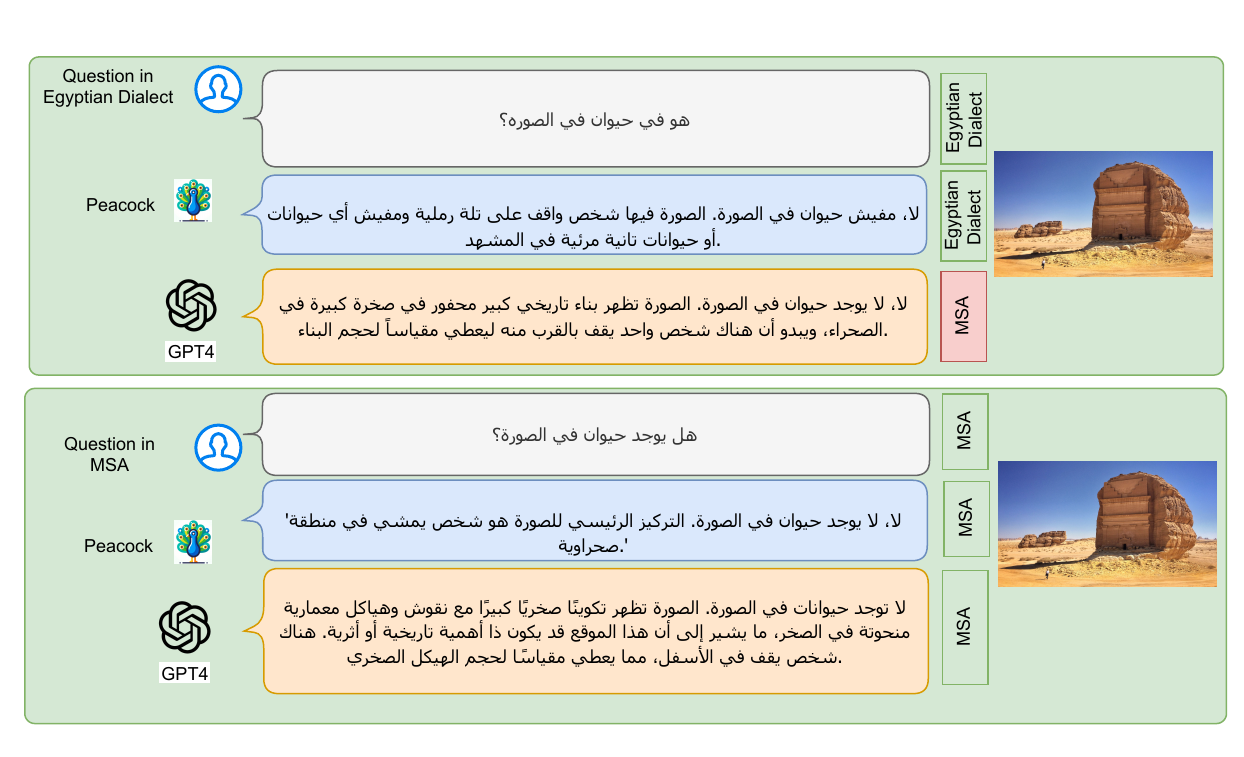}
\centering
\caption{A demonstration of \ours model's advantage over GPT-4V in responding to queries in the Egyptian dialect, with the former responding in the dialect and the latter in MSA. A secondary example shows both models replying in MSA when queried in MSA.} 
\label{fig:egy_dialct_case}
\end{figure*}

Figure \ref{fig:egy_translation} displays three randomly selected examples of question-answer pairs translated from MSA into the Egyptian dialect.
Figure \ref{fig:egy_models_answers} presents four examples—three from \benchmark and one from LLaVA-Bench—along with the correct MSA answers and responses from two models: \colorbox{blue!10}{InstructBlip-\jasmine} and \colorbox{yellow!10}{InstructBlip-AceGPT}, demonstrating the \ours model's proficiency in responding in the Egyptian dialect. In Figure \ref{fig:egy_dialct_case}, we highlight the superiority of \ours over GPT-4 when posed with questions in the Egyptian dialect, where GPT-4 responds in MSA while \ours provides answers in the Egyptian dialect. Conversely, when both models are queried in MSA, they reply in MSA.

\end{document}